\documentclass{nature} %

\newif\ifdraft
\draftfalse

\usepackage[T1]{fontenc}
\usepackage[utf8]{inputenc}
\usepackage{textgreek}

\usepackage{helvet}    %
\usepackage{sectsty}   %
\allsectionsfont{\sffamily}  %

\usepackage{newtxtext,newtxmath}    %
\usepackage{amsmath, amssymb, bm} %
\usepackage{amsthm}               %
\usepackage{graphicx}             %
\usepackage{xspace}               %
\usepackage{booktabs}             %
\usepackage{longtable}            %
\usepackage{multirow}
\usepackage{pdflscape}            %
\usepackage[left=1in,right=1in,bottom=1.1in,top=1in]{geometry}
\usepackage{enumitem}             %
\usepackage[table,dvipsnames]{xcolor}         %
\usepackage{microtype}            %
\usepackage{siunitx}              %
\usepackage{textcomp}
\usepackage[normalem]{ulem}       %
\usepackage[ruled,vlined,linesnumbered]{algorithm2e} %
\usepackage[innercaption]{sidecap} %
\usepackage{lineno}               %
\usepackage{verbatim}             %
\let\spacing\relax
\usepackage{setspace}             %
\usepackage{xr}                   %
\usepackage[colorlinks,citecolor=blue,urlcolor=blue]{hyperref} %
\usepackage{xr-hyper}
\usepackage[capitalise,noabbrev]{cleveref}    %
\usepackage{float}
\usepackage{url}
\usepackage{orcidlink}
\usepackage{docmute}               %
\usepackage{import}               %

\DeclareSIUnit\Molar{M}
\DeclareSIUnit\rpm{rpm}
\DeclareSIUnit\ppm{ppm}

\usepackage[natbib=true, citestyle=numeric-comp,
            bibstyle=nature, sorting=none,
            maxbibnames=6, doi=true, url=false, isbn=false, date=year]{biblatex}

\AtEveryBibitem{%
  \clearfield{note}%
  \clearlist{note}%
  \clearfield{language}%
  \clearlist{language}%
}

\usepackage[font=footnotesize, labelfont={sf,bf}]{caption}
\DeclareCaptionLabelFormat{extended_data}{\textbf{Extended Data #1 #2}}
\DeclareCaptionLabelFormat{extendeddata}{\textsf{\textbf{Extended Data #1 #2}}}
\DeclareCaptionLabelFormat{supplementary}{\textsf{\textbf{Supplementary #1 #2}}}

\newcounter{extfigure}
\newcounter{exttable}

\graphicspath{{./}{FIG/}}

\newcommand{\xhdr}[1]{\vspace{1.2mm}\noindent\textsf{\textbf{{#1.}}}\xspace}

\newcommand{\textsbf}[1]{\textsf{\textbf{#1}}}

\SetKwRepeat{Do}{do}{while}
\let\oldnl\nl
\newcommand{\nonl}{\renewcommand{\nl}{\let\nl\oldnl}}

\theoremstyle{definition}

\usepackage[framemethod=tikz]{mdframed} %
\mdfdefinestyle{promptsllm}{
	innertopmargin=1.2\baselineskip,
	innerbottommargin=0.8\baselineskip,
	roundcorner=5pt,
	nobreak,
	singleextra={%
		\draw(P-|O)node[xshift=1em,anchor=west,fill=Yellow!15,draw,rounded corners=5pt]{%
		  \promptVanillaTitle};
	},
}

\usepackage[most]{tcolorbox}
\newtcolorbox{quotebox}{
  enhanced,
  breakable,
  colback=gray!5,
  colframe=gray!40,
  left=4mm,
  right=2mm,
  boxrule=0pt,
  borderline west={2pt}{0pt}{gray!60},
  sharp corners,
}

\ifdraft
  \newcommand{\draftnote}[1]{{\color{magenta}[Note: #1]}}
  \newcommand{\todo}[1]{{\color{red}\textbf{[TODO: #1]}}}
  \newcommand{\ayush}[1]{{\color{purple}[Ayush: {#1}]}}
  \newcommand{\marinka}[1]{{\color{magenta}[Marinka: {#1}]}}
\else
  \newcommand{\draftnote}[1]{}
  \newcommand{\todo}[1]{}
  \newcommand{\ayush}[1]{}
  \newcommand{\marinka}[1]{}
\fi

\newcommand{\name}{\textsc{ATHENA-R1}\xspace}
 
\newcommand{\toolgen}{\textsc{ToolGen}\xspace} 
\newcommand{\questgen}{\textsc{QuestionGen}\xspace} 
\newcommand{\tracegen}{\textsc{TraceGen}\xspace} 
\newcommand{\toolrag}{\textsc{ToolRAG} model\xspace} 
\newcommand{\toolbox}{tool library\xspace}
\newcommand{\trainset}{\textsc{ATHENA-R1-Instruct}\xspace}
\newcommand{\onlineref}[1]{Methods~\ref{#1}\xspace}

\newcommand{\extfigref}[1]{Extended Data Figure~\ref{#1}}

\newcommand{\supfigref}[1]{Supplementary Figure~\ref{#1}}
\newcommand{\suptabref}[1]{Supplementary Table~\ref{#1}}

\newcounter{sinote}
\renewcommand{\thesinote}{\arabic{sinote}}

\title{\begin{center}
An AI agent for treatment reasoning \\ over a biomedical tool universe
\end{center}\vspace{-10mm}}    

\author  
{\small\begin{center}   
Shanghua~Gao$^{1}$,
Ayush~Noori$^{1,2,3}$~\orcidlink{0000-0003-1420-1236},
Richard~Zhu$^{1}$~\orcidlink{0009-0004-6190-8503},
Curtis~Ginder$^{1,4}$~\orcidlink{0000-0001-8507-9624},
Zhenglun~Kong$^{1}$~\orcidlink{0000-0002-8120-4456},
Xiaorui~Su$^{1}$,
\\ %
Justin~Kauffman$^{5}$~\orcidlink{0009-0004-6371-4198},
Benjamin~S.~Glicksberg$^{5,6,7}$~\orcidlink{0000-0003-4515-8090},
Joshua~Lampert$^{5,6,8}$,
Ankit~Sakhuja$^{5,9,10}$,
Ashwin~Sawant$^{5,9,11}$~\orcidlink{0000-0003-1525-8541},
ATHENA-R1~Evaluation~Consortium$^{12}$,
David~A.~Clifton$^{2,13}$~\orcidlink{0000-0002-9848-8555},
Noa~Dagan$^{3,14,15}$~\orcidlink{0000-0001-8811-7825},
Ran~Balicer$^{3,14,16}$~\orcidlink{0000-0002-7783-6362},
Marinka~Zitnik$^{1,3,17,18,19,\dagger}$~\orcidlink{0000-0001-8530-7228} \\[2mm]  
\footnotesize{$^{1}$Department of Biomedical Informatics, Harvard Medical School, Boston, MA} \\
\footnotesize{$^{2}$Department of Engineering Science, University of Oxford, Oxford, UK} \\
\footnotesize{$^{3}$The Ivan and Francesca Berkowitz Family Living Laboratory Collaboration at \\ Harvard Medical School and Clalit Research Institute, Boston, MA, USA} \\
\footnotesize{$^{4}$Cardiovascular Division, Department of Medicine, Brigham and Women’s Hospital, \\ Harvard Medical School, Boston, MA} \\
\footnotesize{$^{5}$The Windreich Department of Artificial Intelligence and Human Health, \\Icahn School of Medicine at Mount Sinai, New York, NY, USA} \\
\footnotesize{$^{6}$The Hasso Plattner Institute for Digital Health at Mount Sinai, Icahn School of Medicine\\ at Mount Sinai and Mount Sinai Health System, New York City, NY, USA} \\
\footnotesize{$^{7}$Mindich Child Health and Development Institute and the Departments of Pediatrics\\ and Genetics \& Genomic Sciences, Icahn School of Medicine at Mount Sinai, New York, NY, USA} \\
\footnotesize{$^{8}$Mount Sinai Fuster Heart Hospital, Icahn School of Medicine at Mount Sinai, New York, NY, USA} \\
\footnotesize{$^{9}$Mount Sinai AI Assurance Lab, Mount Sinai Health System, New York, NY, USA} \\ 
\footnotesize{$^{10}$Institute for Critical Care Medicine, Icahn School of Medicine at Mount Sinai, New York, NY, USA} \\ 
\footnotesize{$^{11}$Department of Medicine, Icahn School of Medicine at Mount Sinai, New York, NY, USA} \\ 
\footnotesize{$^{12}$ATHENA-R1 Evaluation Group (the list of members and their affiliations appears in the Supplementary Information)} \\
\footnotesize{$^{13}$}Oxford Suzhou Centre for Advanced Research, University of Oxford, Suzhou, Jiangsu, China\\
\footnotesize{$^{14}$Clalit Research Institute, Innovation Division, Clalit Health Services, Ramat Gan, Israel} \\
\footnotesize{$^{15}$Faculty of  Computer and Information Science, Ben Gurion University of the Negev, Be'er Sheva, Israel} \\
\footnotesize{$^{16}$Faculty of Health Sciences, School of Public Health, Ben Gurion University of the Negev, Be'er Sheva, Israel} \\
\footnotesize{$^{17}$Kempner Institute for the Study of Natural and Artificial Intelligence, Harvard University, Cambridge, MA} \\
\footnotesize{$^{18}$Broad Institute of MIT and Harvard, Cambridge, MA} \\
\footnotesize{$^{19}$Harvard Data Science Initiative, Cambridge, MA} \\ [1mm]
\footnotesize{$^\dagger$Correspondence: \href{mailto:marinka@hms.harvard.edu}{marinka@hms.harvard.edu}}
\end{center}
}

\makeatletter
\def\xrefdef#1#2{\global\expandafter\def\csname r@#1\endcsname{#2}}
\xrefdef{sec:note1}{{1}{1}{Architecture, skills, and \name inference}{section.1}{}}
\xrefdef{sec:note1@cref}{{[section][1][]1}{[1][1][]1}{}{}{}}
\xrefdef{sec:skill_txagent}{{1.2}{2}{Skills of \name }{subsection.1.2}{}}
\xrefdef{sec:skill_txagent@cref}{{[subsection][2][1]1.2}{[1][2][]2}{}{}{}}
\xrefdef{alg:txagent_inference}{{1}{5}{Skills of \name }{algocf.1}{}}
\xrefdef{alg:txagent_inference@cref}{{[algorithm][1][]1}{[1][5][]5}{}{}{}}
\xrefdef{sec:cap_txagent}{{1.3}{5}{Core agentic abilities of \name }{subsection.1.3}{}}
\xrefdef{sec:cap_txagent@cref}{{[subsection][3][1]1.3}{[1][5][]5}{}{}{}}
\xrefdef{sec:note2}{{2}{7}{Biomedical tool library}{section.2}{}}
\xrefdef{sec:note2@cref}{{[section][2][]2}{[1][6][]7}{}{}{}}
\xrefdef{sec:data_gen_method}{{2.1}{7}{\toolgen : a multi-agent system for constructing tools}{subsection.2.1}{}}
\xrefdef{sec:data_gen_method@cref}{{[subsection][1][2]2.1}{[1][7][]7}{}{}{}}
\xrefdef{sec:note3}{{3}{8}{Self-learning in \name }{section.3}{}}
\xrefdef{sec:note3@cref}{{[section][3][]3}{[1][8][]8}{}{}{}}
\xrefdef{sec:train_set}{{3}{8}{Self-learning in \name }{section.3}{}}
\xrefdef{sec:train_set@cref}{{[section][3][]3}{[1][8][]8}{}{}{}}
\xrefdef{label:data_source}{{3.1}{8}{\trainset Data Sources}{subsection.3.1}{}}
\xrefdef{label:data_source@cref}{{[subsection][1][3]3.1}{[1][8][]8}{}{}{}}
\xrefdef{sec:question_gen}{{3.3}{10}{\questgen multi-agent system for question construction}{subsection.3.3}{}}
\xrefdef{sec:question_gen@cref}{{[subsection][3][3]3.3}{[1][10][]10}{}{}{}}
\xrefdef{sec:trace_gen}{{3.4}{12}{Reasoning trace generation}{subsection.3.4}{}}
\xrefdef{sec:trace_gen@cref}{{[subsection][4][3]3.4}{[1][12][]12}{}{}{}}
\xrefdef{alg:solver_trace}{{2}{15}{Reasoning trace generation}{algocf.2}{}}
\xrefdef{alg:solver_trace@cref}{{[algorithm][2][]2}{[1][14][]15}{}{}{}}
\xrefdef{sec:note4}{{3.6}{16}{\name level-1 training: Data preparation and augmentation}{subsection.3.6}{}}
\xrefdef{sec:note4@cref}{{[subsection][6][3]3.6}{[1][16][]16}{}{}{}}
\xrefdef{sec:training_txagent}{{3.6}{16}{\name level-1 training: Data preparation and augmentation}{subsection.3.6}{}}
\xrefdef{sec:training_txagent@cref}{{[subsection][6][3]3.6}{[1][16][]16}{}{}{}}
\xrefdef{sec:dataset_convert}{{3.6}{16}{\name level-1 training: Data preparation and augmentation}{subsection.3.6}{}}
\xrefdef{sec:dataset_convert@cref}{{[subsection][6][3]3.6}{[1][16][]16}{}{}{}}
\xrefdef{sec:training_design}{{3.7}{19}{\name level-1 training: Training design}{subsection.3.7}{}}
\xrefdef{sec:training_design@cref}{{[subsection][7][3]3.7}{[1][18][]19}{}{}{}}
\xrefdef{sec:rl_training}{{3.8}{20}{\name level-2 training: Reinforcement learning}{subsection.3.8}{}}
\xrefdef{sec:rl_training@cref}{{[subsection][8][3]3.8}{[1][20][]20}{}{}{}}
\xrefdef{tab:reward_components}{{1}{22}{Scientific feedback reward: 12 checks grouped into six dimensions. Checks marked with $\dagger $ are conditionally scaled to 80\% of their listed weight when the evidence-gathering check is not satisfied}{table.caption.3}{}}
\xrefdef{tab:reward_components@cref}{{[table][1][]1}{[1][21][]22}{}{}{}}
\xrefdef{alg:grpo_training}{{3}{23}{\name level-2 training: Reinforcement learning}{algocf.3}{}}
\xrefdef{alg:grpo_training@cref}{{[algorithm][3][]3}{[1][22][]23}{}{}{}}
\xrefdef{sec:note5}{{4}{23}{Model benchmarking}{section.4}{}}
\xrefdef{sec:note5@cref}{{[section][4][]4}{[1][23][]23}{}{}{}}
\xrefdef{sec:benchmark_details}{{4}{23}{Model benchmarking}{section.4}{}}
\xrefdef{sec:benchmark_details@cref}{{[section][4][]4}{[1][23][]23}{}{}{}}
\xrefdef{sec:eval_strategy}{{4.2}{26}{Evaluation strategy}{subsection.4.2}{}}
\xrefdef{sec:eval_strategy@cref}{{[subsection][2][4]4.2}{[1][26][]26}{}{}{}}
\xrefdef{sec:human_eval}{{5}{26}{Real-world evaluation: Rare disease treatment reasoning}{section.5}{}}
\xrefdef{sec:human_eval@cref}{{[section][5][]5}{[1][26][]26}{}{}{}}
\xrefdef{sec:mount_sinai}{{6}{30}{Real-world evaluation: Complex clinical cases}{section.6}{}}
\xrefdef{sec:mount_sinai@cref}{{[section][6][]6}{[1][30][]30}{}{}{}}
\xrefdef{sec:ehr_eval}{{7}{31}{Real-world evaluation: Population-scale EHR analyses}{section.7}{}}
\xrefdef{sec:ehr_eval@cref}{{[section][7][]7}{[1][31][]31}{}{}{}}
\xrefdef{sec:note1_si}{{1}{4}{Supplementary Note \thesinote {} $|$ Generalization across drug name variants and descriptions}{section*.1}{}}
\xrefdef{sec:note1_si@cref}{{[sinote][1][]1}{[1][4][]4}{}{}{}}
\xrefdef{sec:note3_si}{{2}{5}{Supplementary Note \thesinote {} $|$ Multi-step reasoning ablations and tool comparisons}{section*.2}{}}
\xrefdef{sec:note3_si@cref}{{[sinote][2][]2}{[1][4][]5}{}{}{}}
\xrefdef{sec:no_thought_inference}{{2.1}{5}{The role of reasoning thoughts in model performance}{subsection.2.1}{}}
\xrefdef{sec:no_thought_inference@cref}{{[subsection][1][2]2.1}{[1][5][]5}{}{}{}}
\xrefdef{alg:txagent_inference_nothought}{{1}{6}{The role of reasoning thoughts in model performance}{algocf.1}{}}
\xrefdef{alg:txagent_inference_nothought@cref}{{[algorithm][1][]1}{[1][5][]6}{}{}{}}
\xrefdef{sec:trace_length}{{2.2}{6}{Longer training traces improve complex reasoning}{subsection.2.2}{}}
\xrefdef{sec:trace_length@cref}{{[subsection][2][2]2.2}{[1][6][]6}{}{}{}}
\xrefdef{sec:inference_step_limit}{{2.3}{6}{Longer inference enables evidence accumulation}{subsection.2.3}{}}
\xrefdef{sec:inference_step_limit@cref}{{[subsection][3][2]2.3}{[1][6][]6}{}{}{}}
\xrefdef{sec:step_tool_usage}{{2.4}{7}{Reasoning step and tool-call usage across benchmarks}{subsection.2.4}{}}
\xrefdef{sec:step_tool_usage@cref}{{[subsection][4][2]2.4}{[1][7][]7}{}{}{}}
\xrefdef{sec:llm_as_tool}{{2.5}{7}{Tool execution outperforms LLM-simulated tools}{subsection.2.5}{}}
\xrefdef{sec:llm_as_tool@cref}{{[subsection][5][2]2.5}{[1][7][]7}{}{}{}}
\xrefdef{sec:tool_scaling}{{2.6}{8}{Expanding the tool library improves performance}{subsection.2.6}{}}
\xrefdef{sec:tool_scaling@cref}{{[subsection][6][2]2.6}{[1][7][]8}{}{}{}}
\xrefdef{sec:gpt5_tools}{{2.7}{8}{Trained tool-use reasoning outperforms tool access alone}{subsection.2.7}{}}
\xrefdef{sec:gpt5_tools@cref}{{[subsection][7][2]2.7}{[1][8][]8}{}{}{}}
\xrefdef{sec:note4_si}{{3}{9}{Supplementary Note \thesinote {} $|$ System and agent prompt designs}{section*.3}{}}
\xrefdef{sec:note4_si@cref}{{[sinote][3][]3}{[1][9][]9}{}{}{}}
\xrefdef{sec:note5_si}{{4}{16}{Supplementary Note \thesinote {} $|$ Additional details on multi-institu\-tional human evaluation}{section*.4}{}}
\xrefdef{sec:note5_si@cref}{{[sinote][4][]4}{[1][15][]16}{}{}{}}
\xrefdef{sec:eval_criteria}{{4.1}{16}{Definition of evaluation criteria and scoring rubrics}{subsection.4.1}{}}
\xrefdef{sec:eval_criteria@cref}{{[subsection][1][4]4.1}{[1][15][]16}{}{}{}}
\xrefdef{tab:pairwise_extended}{{1}{20}{Distribution of pairwise preferences across all 110 evaluated responses. Values are reported as counts, with percentages shown in parentheses}{table.caption.5}{}}
\xrefdef{tab:pairwise_extended@cref}{{[table][1][]1}{[1][19][]20}{}{}{}}
\xrefdef{tab:absolute_extended}{{2}{20}{Mean absolute ratings (1--5 Likert scale) per criterion for \name and reference models, computed as per-response averages across all valid ratings. The number of valid ratings per criterion varies (45--110) because evaluators could select ``Unable to Judge.'' Reference models include Qwen3 ($n=100$), o3-mini ($n=3$), Gemini-2.0-Flash ($n=3$), DeepSeek-R1 ($n=2$), DeepSeek-R1-Distill-Llama-8B ($n=1$), and Llama-3.1-8B-Instruct ($n=1$)}{table.caption.6}{}}
\xrefdef{tab:absolute_extended@cref}{{[table][2][]2}{[1][19][]20}{}{}{}}
\xrefdef{tab:stat_tests}{{3}{21}{Per-criterion statistical tests comparing \name against reference models on 110 expert-evaluated responses. Two complementary tests are reported: a \textbf {one-sided binomial test} on pairwise preferences, assessing whether evaluators select \name more often than chance ($p_0 = 0.5$); and a \textbf {one-sided Wilcoxon signed-rank test} on paired absolute Likert ratings (\name minus reference), assessing whether \name scores are systematically higher. \textit {Win rate} is the fraction of decisive comparisons favoring \name , shown as $k/n$ (count) with the percentage in parentheses, after excluding ``Both are equally good'' and ``Neither did well.'' \textit {$n$} under the Wilcoxon test is the number of question-level paired ratings, after excluding ``Unable to Judge'' responses pairwise; it differs from the binomial denominator because the two exclusion rules apply to different response options. Note that the two $n$ values come from different sub-samples and are not directly comparable. \textit {$P$} values from both tests are shown uncorrected; all eight criteria satisfy $P < 5 \times 10^{-5}$ under both tests. \textit {Effect size $r$} is the matched-pairs rank-biserial correlation for the Wilcoxon test, computed as $r = (W^+ - W^-)/(W^+ + W^-)$~\cite {kerby2014simple}, with $r \in [-1, 1]$; $|r| > 0.5$ denotes a large effect, and all eight criteria show $r \geq 0.71$}{table.caption.7}{}}
\xrefdef{tab:stat_tests@cref}{{[table][3][]3}{[1][20][]21}{}{}{}}
\xrefdef{sec:note7_si}{{5}{22}{Supplementary Note \thesinote {} $|$ Clinical case-based expert review: Full questions, reasoning traces, and final answers}{section*.8}{}}
\xrefdef{sec:note7_si@cref}{{[sinote][5][]5}{[1][22][]22}{}{}{}}
\xrefdef{sec:si_ae_generation}{{6}{32}{Supplementary Note \thesinote {} $|$ Generation, scoring, and ranking of treatment-associated risk hypotheses}{section*.9}{}}
\xrefdef{sec:si_ae_generation@cref}{{[sinote][6][]6}{[1][32][]32}{}{}{}}
\xrefdef{sec:note6_si}{{7}{34}{Supplementary Note \thesinote {} $|$ Potential sources of residual confounding in health records analyses}{section*.10}{}}
\xrefdef{sec:note6_si@cref}{{[sinote][7][]7}{[1][34][]34}{}{}{}}
\xrefdef{sec:note8_si}{{8}{35}{Supplementary Note \thesinote {} $|$ Sampling variability}{section*.12}{}}
\xrefdef{sec:note8_si@cref}{{[sinote][8][]8}{[1][35][]35}{}{}{}}
\xrefdef{fig:multi-agent-system}{{1}{36}{\textsf {\textbf {Self-learning level 1: multi-agent systems construct the \trainset dataset.}} The first level of \name 's self-learning constructs the tools and training data via three multi-agent systems, whose outputs form \trainset for supervised fine-tuning. \textbf {a}) \trainset is a diverse synthetic multi-step reasoning and massive tool call training dataset anchored in biomedical knowledge. Three datasets are built by the auxiliary agent systems: a tooling dataset (augmented versions of 212 tools from \toolbox ), a treatment task dataset (85,340 tasks generated by \questgen ), and a reasoning trace dataset (85,340 traces comprising 177,626 reasoning steps and 281,695 tool calls, generated by \tracegen ). Processing these three datasets yields \trainset with 378,027 instruction-tuning samples. \textbf {b}) \toolgen : a tool generation multi-agent system that transforms APIs into 212 agent-compatible tools aggregated into \toolbox . \textbf {c}) \questgen : a question generation multi-agent system that extracts critical information from documents (e.g., FDA drug documentation) and generates relevant treatment tasks. \textbf {d}) \tracegen : a reasoning trace generation multi-agent system in which a \textsc {Helper} agent (with access to the ground-truth answer) and a \textsc {Tool Provider} module assist the \textsc {Solver} agent in generating step-by-step reasoning and tool calls}{figure.caption.13}{}}
\xrefdef{fig:multi-agent-system@cref}{{[figure][1][]1}{[1][36][]36}{}{}{}}
\xrefdef{fig:rl-system}{{2}{37}{\textsf {\textbf {Self-learning level 2: reinforcement learning with scientific feedback.}} The second level refines \name 's policy on its own rollouts, with rewards computed by rule-based scientific feedback. \textbf {a}) RL training loop. For each training prompt, \name samples $n{=}5$ candidate rollouts, each scored on six rule-based dimensions by a composite scientific-feedback reward; the policy is then updated by group relative policy optimization (GRPO). The loop iterates over gradient steps. \textbf {b}) The six rule-based dimensions aggregate 12 individual checks: answer correctness, output-format validity, evidence gathering, multi-step reasoning, tool-argument grounding, and reasoning non-redundancy. Full per-check specification and weights are given in Table~\ref {tab:reward_components}}{figure.caption.14}{}}
\xrefdef{fig:rl-system@cref}{{[figure][2][]2}{[1][36][]37}{}{}{}}
\xrefdef{fig:avg_infer_step}{{3}{38}{\textsf {\textbf {Multi-step reasoning ablations, tool comparisons, and inference statistics across benchmarks.}} \textbf {a}) Explicit thought generation is fundamental to reasoning in \name . Removing thought generation reduces accuracy by 22.3\% on DrugPC and 21.9\% on TreatmentPC multiple-choice benchmarks. \textbf {b}) Long multi-step traces in training data enhance \name 's ability to handle complex tasks. When training data is limited to single-step reasoning traces, accuracy drops from 93.8\% to 71.6\% on DrugPC and from 86.8\% to 66.9\% on TreatmentPC, with the larger decline on TreatmentPC indicating that treatment recommendation requires deeper multi-step reasoning. \textbf {c}) Longer inference traces enhance model performance. Restricting \name to a single inference step reduces accuracy to 73.5\%, 13.3\% below unrestricted reasoning. Performance improves with additional steps, with diminishing gains beyond five steps}{figure.caption.15}{}}
\xrefdef{fig:avg_infer_step@cref}{{[figure][3][]3}{[1][38][]38}{}{}{}}
\xrefdef{fig:descriptionpc}{{4}{40}{\textsf {\textbf {Performance of \name on drug name variant and description benchmarks.}} \textbf {a}) \name surpasses both native and tool-use LLMs on the DrugPC benchmark, as well as its Brand and Generic variants, where drug names are replaced with their brand and generic counterparts. Additionally, \name demonstrates minimal variance when handling drug names with different representations. \textbf {b}) \name surpasses LLM in a two-step evaluation on the DescriptionPC benchmark, where drug names are replaced with their descriptions, including indications, mechanisms of action, contraindications, and interactions. In this evaluation, the first step involves identifying the correct drug name based on its description, followed by answering the question using the correctly identified drug name. All \name results in this figure use the Llama-3.1-8B-based ATHENA-R1-Llama8B configuration}{figure.caption.17}{}}
\xrefdef{fig:descriptionpc@cref}{{[figure][4][]4}{[1][40][]40}{}{}{}}
\xrefdef{fig:comparison_deepseek}{{5}{41}{\textsf {\textbf {Comparison of \name and DeepSeek-R1 on a pediatric treatment-safety question.}} For a 12-year-old patient with alopecia areata and heterozygous familial hypercholesterolemia, the task is to identify which of two candidate medications should be avoided. DeepSeek-R1 reasons from internal knowledge, incorrectly concluding that intralesional triamcinolone acetonide is safe in children and that both medications are acceptable, selecting the wrong option. In contrast, \name retrieves the FDA label through tool calls and identifies documented pediatric risks of triamcinolone acetonide, including HPA-axis suppression, Cushing's syndrome and intracranial hypertension, while confirming that ZOCOR (simvastatin) is established as safe in patients aged 10 and older. Grounding its reasoning in retrieved labels allows \name to reach the correct, verifiable conclusion where DeepSeek-R1 hallucinates}{figure.caption.18}{}}
\xrefdef{fig:comparison_deepseek@cref}{{[figure][5][]5}{[1][41][]41}{}{}{}}
\xrefdef{fig:reproducibility}{{6}{42}{\textsf {\textbf {Sampling variability of \name accuracy across five independent rollouts.}} Bars show the mean accuracy across $n{=}5$ independent rollouts of \name ; error bars are sample standard deviation (SD). Two answer-extraction protocols are reported per benchmark: \emph {self as judge}, in which \name selects the correct option letter from its own free-form answer, and \emph {GPT-5 as judge}, in which an independent GPT-5 instance maps \name 's free-form answer to an option letter. \textbf {a}, Aggregate accuracy on TreatmentPC (456 patient-specific multiple-choice treatment tasks; left) and DrugPC (3{,}168 FDA-labeling questions; right). Main-text Figure~\ref {fig:fig2} reports one independent rollout for each benchmark and protocol; the corresponding main-text values $94.7\%$ (DrugPC, GPT-5 as judge), $82.9\%$ (TreatmentPC, GPT-5 as judge) and $74.8\%$ (TreatmentPC, self as judge) all fall within mean $\pm 1\sigma $ of the rollout distributions shown here. \textbf {b}, DrugPC per-task sampling variability across the 11 task categories defined in Extended Data Table~\ref {table:fda_11tasks}, ordered by descending per-task sample size $n_q$. Protocol is reported in Supplementary Note~\ref {sec:note8_si}}{figure.caption.19}{}}
\xrefdef{fig:reproducibility@cref}{{[figure][6][]6}{[1][42][]42}{}{}{}}
\xrefdef{fig:extend_4abilities}{{1}{43}{\textsf {\textbf {Key features of \name .}} \textbf {a}) Knowledge grounding using tool calls, where \name uses tools to obtain verified knowledge and provides outputs based on it. \textbf {b}) Goal-oriented tool selection, where \name proactively requests tools from \toolbox using the \toolrag model and selects and applies the most suitable tool from the available candidates. \textbf {c}) Problem solving with multi-step reasoning, where \name manages complex tasks or unexpected responses from tools through multiple iterations of thought and tool calls. \textbf {d})~Leveraging constantly updated knowledge bases, where \name accesses continuously updated databases via tools to handle problems that go beyond \name 's intrinsic knowledge}{figure.caption.20}{}}
\xrefdef{fig:extend_4abilities@cref}{{[figure][1][]1}{[1][43][]43}{}{}{}}
\xrefdef{fig:tool_description}{{2}{44}{\textsf {\textbf {Tool specification examples in \toolbox .}} Each specification includes a tool description, which serves as a reference for \name 's tool calls, and a mapping rule that translates tool calls into API requests. The tool description outlines the tool's name, purpose, and the arguments it accepts, including details such as each argument's name, purpose, data type, and whether it is mandatory. \textbf {a}) Tool description for the tool from openFDA. \textbf {b}) Tool description for the tool from Open Targets. \textbf {c}) Mapping between tools in \name and external APIs from openFDA~\cite {kass2016openfda}. \textbf {d}) Mapping between tools in \name and external APIs from Open Targets~\cite {buniello2025open}}{figure.caption.21}{}}
\xrefdef{fig:tool_description@cref}{{[figure][2][]2}{[1][44][]44}{}{}{}}
\xrefdef{fig:tool_dist}{{3}{45}{\textsf {\textbf {Categories of biomedical tools in \toolbox .}} The \toolbox contains 212 biomedical tools and includes the following categories: adverse events, risks, safety; addiction and abuse; drug usage in patient populations; drug administration and handling; pharmacology; drug use, mechanism, composition; id and labeling tools; general clinical annotations; clinical laboratory info; general info for patients and relatives; disease, phenotype, target, drug links; biological annotation tools; publications; search; target characterization}{figure.caption.22}{}}
\xrefdef{fig:tool_dist@cref}{{[figure][3][]3}{[1][44][]45}{}{}{}}
\xrefdef{fig:si_case4_case5_ratings}{{4}{46}{\textsf {\textbf {Expert absolute ratings for Cases~4 and 5 in the clinical expert review of complex, real-world treatment decisions.}} Physicians at the same institution independently rated \name 's responses on the eight-criterion 1--5 Likert rubric (Supplementary Note~\ref {sec:note7_si}). Bars show the mean across reviewers; dots show individual reviewer scores with small vertical jitter applied for visibility. Bar colors group criteria by category (clinical accuracy, reasoning, task performance, safety). \textbf {a}) Case~4: perioperative pain management and polypharmacy in a hip-fracture patient. Three physicians each scored all eight criteria; individual scores ranged only from 2 to 3, with no criterion receiving a 4 or 5 from any reviewer. On four criteria (task success, cognitive traceability, accuracy of content, clinical relevance) all three reviewers independently assigned a score of 3, and the three dots coincide near $x{=}3$; on the remaining four criteria one reviewer assigned 2 while the other two assigned 3. The consensus at score~3 corresponds to the rubric anchor ``addressed the task but with notable limitations'' and indicates that \name 's response on this case was judged satisfactory but not strong across all eight criteria. \textbf {b}) Case~5: empirical antibiotic selection in a preterm infant with suspected necrotizing enterocolitis. Of the three physician reviewers, one left the scoring sheet blank and another marked every criterion as ``NA'' (not applicable); only the third reviewer provided numeric scores. Bars and dots therefore reflect a single-reviewer rating ($n{=}1$) rather than a three-reviewer mean, and this panel is included for completeness of the five-case vignette set rather than as a basis for quantitative comparison. The complete reasoning trace and the reviewer's written rationale are reported in Supplementary Note~\ref {sec:note7_si}, Case~5}{figure.caption.23}{}}
\xrefdef{fig:si_case4_case5_ratings@cref}{{[figure][4][]4}{[1][46][]46}{}{}{}}
\xrefdef{table:fda_11tasks}{{1}{47}{\textsf {\textbf {Benchmark datasets derived from FDA drugs newly approved in 2024.}} New FDA approved drugs were chosen to minimize information leakage from LLM pre-training. Questions and answers were reviewed by human curators to exclude non-biomedical items}{table.1}{}}
\xrefdef{table:fda_11tasks@cref}{{[table][1][]1}{[1][47][]47}{}{}{}}
\xrefdef{tab:notation}{{1}{48}{\textsf {\textbf {Notation used in methodological derivations.}} Symbols for questions, answers, reasoning steps, tool calls and intermediate quantities}{table.caption.24}{}}
\xrefdef{tab:notation@cref}{{[table][1][]1}{[1][48][]48}{}{}{}}
\xrefdef{tab:example_question_type}{{2}{49}{\textsf {\textbf {Example treatment questions, options and answers in the open-ended and multiple-choice evaluation settings.}} The same clinical scenario is shown under both protocols: the multiple-choice setting requires selecting the correct option letter, while the open-ended setting requires generating a free-form answer with rationale}{table.caption.25}{}}
\xrefdef{tab:example_question_type@cref}{{[table][2][]2}{[1][49][]49}{}{}{}}
\xrefdef{tab:tool_list}{{3}{50}{\textsf {\textbf {Biomedical tools in the tool library.}} The tool library contains 212 biomedical tools constructed by \toolgen from three API sources: openFDA (FDA drug labeling), Open Targets (disease--target--drug associations), and Human Phenotype Ontology (phenotype--disease associations). Each tool is defined by a name, a natural-language description, and a structured parameter schema used by \name to call external APIs during multi-step reasoning}{table.3}{}}
\xrefdef{tab:tool_list@cref}{{[table][3][]3}{[1][50][]50}{}{}{}}
\xrefdef{tab:thera_consortium}{{4}{68}{\textsf {\textbf {\textsc {ATHENA-R1} Evaluation Consortium: rare disease experts contributing to the blinded treatment reasoning evaluation.}} The consortium comprises 29 experts representing 28 patient-led and clinical organizations focused on rare diseases, recruited through the Chan Zuckerberg Initiative Rare As One network and collaborating institutions. Each expert provided disease-specific context used by \questgen to construct treatment cases for the blinded, arena-based evaluation reported in Figure~\ref {fig:fig3} and Supplementary Note~\ref {sec:note5_si}. Two experts (Ali Rosenberg, Dianne Mitchell) hold appointments at two organizations}{table.4}{}}
\xrefdef{tab:thera_consortium@cref}{{[table][4][]4}{[1][68][]68}{}{}{}}
\makeatother

\begin{document}
\maketitle

\begin{center}\small
\href{https://athena.openscientist.ai}{Project page}\quad\textcolor{gray!70}{\textbar}\quad
\href{https://github.com/mims-harvard/ATHENA}{Code}\quad\textcolor{gray!70}{\textbar}\quad
\href{https://athena.openscientist.ai/pdf/ATHENA-R1_Online_Methods.pdf}{Online Methods}\quad\textcolor{gray!70}{\textbar}\quad
\href{https://athena.openscientist.ai/pdf/ATHENA-R1_Supplementary_Information.pdf}{Supplementary Information}
\end{center}

\vspace{1em}
\begin{spacing}{1}
\small
\begin{abstract}
\section*{Abstract}

Treatment reasoning underpins every therapeutic decision in medicine, requiring the integration of disease context, comorbidities, concurrent medications, contraindications, and evolving biomedical knowledge to arrive at a therapy appropriate for an individual patient.
This process is inherently iterative, as candidate treatments must be evaluated against multiple constraints, revised as new evidence emerges, and grounded in sources that can be inspected and verified.
Here we introduce \name, an AI agent for treatment reasoning across all FDA approved drugs since 1939, trained through reinforcement learning over a universe of 212 biomedical tools. 
At each reasoning step, \name identifies missing information, selects and executes relevant tools, and incorporates the retrieved evidence before proceeding. To train \name without human-annotated reasoning traces, we develop a two-level self-learning framework in which multi-agent systems construct tools, treatment tasks, and reasoning trajectories for supervised fine-tuning, followed by reinforcement learning with scientific feedback using rewards for reasoning quality, including evidence gathering, grounded tool use, and logical non-redundancy, to refine \name's evidence-seeking strategy.
Across five benchmarks spanning 3,168 drug reasoning tasks and 456 patient treatment cases, \name outperforms  language models and tool-use systems. \name achieves 94.7\% accuracy on open-ended drug reasoning and 82.9\% accuracy on treatment reasoning, exceeding GPT-5 by 17.8 and 10.7 percentage points, respectively. 
In blinded evaluations involving experts from 28 rare disease organizations, \name is preferred over reference models across all evaluation criteria. 
Physicians rated \name favorably on complex hospitalized patient cases spanning cardiovascular management and infectious disease.
Adverse event hypotheses generated by \name were tested in electronic health records from 5.4 million patients, with predicted associations reaching adjusted odds ratios of 1.48--1.84 and negative controls showing no elevation.
These results establish that treatment reasoning, long considered difficult for AI because it requires knowing what evidence to seek before a conclusion can be formed, can be reframed as a learnable process of iterative evidence gathering, and that reinforcement learning can train AI to perform it.

\end{abstract}
\end{spacing}

\begin{spacing}{1.2}

\section*{Main}

Treatment reasoning is among the most demanding tasks in medicine, where selecting a therapy for an individual patient requires weighing disease context, patient characteristics, concurrent medications, safety constraints, and evolving evidence~\cite{hamburg2010path,topol2019high}. Unlike fact retrieval or pattern recognition, treatment reasoning is an iterative process in which candidate strategies must be gathered, evaluated against multiple constraints, and revised until the evidence supports a decision.

Large language models (LLMs) access medical knowledge through pretraining~\cite{dubey2024llama}, biomedical alignment~\cite{singhal2023large,singhal2025toward,chen2023meditron,mcduff2025towards}, and agentic frameworks~\cite{gao2024agent,tu2025towards}. These models generate fluent responses and capture broad clinical patterns, but they rely on parametric knowledge stored in model weights, lack access to updated and vetted medical information, and can produce recommendations that fail to account for relevant contraindications, interactions or patient-specific constraints. Retrieval augmented generation~\cite{gao2023retrieval} and tool-augmented LLMs~\cite{berkeley-function-calling-leaderboard,watt-tool-8B,liu2024toolace} can give LLMs access to information outside their model weights, such as medical documents, biomedical databases and software tools, at inference time. However, access to biomedical tools does not by itself produce treatment reasoning. A model must determine what evidence is needed, select the appropriate source, interpret the result in the context of accumulated evidence, and revise its analysis when evidence is incomplete or conflicting. This capacity cannot be assumed from tool access alone and must be learned.

We introduce \name, an AI agent for treatment reasoning that combines multi-step analysis with direct access to medical evidence. Rather than producing answers in a single step, \name determines what information is needed, retrieves relevant evidence, and uses that evidence to update its analysis. In each reasoning step, \name selects from a library of 212 biomedical tools, retrieves information about drugs, diseases and patient populations, interprets the returned evidence, and incorporates it into subsequent reasoning steps. This allows \name to evaluate candidate treatments through iterative evidence gathering and analysis rather than relying solely on knowledge stored in model weights.

Generating multi-step treatment-reasoning traces at the scale and diversity required for training cannot feasibly be done by human annotators, as each trace must specify what evidence is needed, which tools to call and how retrieved information should be interpreted across hundreds of tools and diverse drug, disease and patient contexts. We therefore train \name through two sequential stages. First, a multi-agent system automatically constructs biomedical tools, treatment tasks and reasoning traces, yielding \trainset, a dataset of 378,027 instruction-tuning samples derived from 85,340 reasoning traces, comprising 177,626 reasoning steps and 281,695 tool calls grounded in all US FDA approved drugs since 1939. After supervised fine-tuning on \trainset, \name is refined through reinforcement learning in a live 212-tool environment, receiving rule-based scientific feedback across six dimensions of reasoning quality, including answer correctness, evidence gathering, multi-step reasoning and tool-use validity.

We evaluate \name on five datasets spanning drug reasoning and patient treatment cases. DrugPC contains 3,168 treatment cases covering 11 treatment tasks, including indications, dosing, safety and pharmacology. BrandPC and GenericPC replace drug names with brand and generic variants, and DescriptionPC replaces names with textual descriptions. TreatmentPC contains 456 treatment cases in which the correct answer depends on patient-specific constraints. Across these datasets, \name consistently outperforms LLMs and tool-use models in open-ended evaluation. On DrugPC, \name achieves 94.7\% accuracy, exceeding GPT-5~\cite{singh2025openai} by 17.8 percentage points and DeepSeek-R1 (671B)~\cite{guo2025deepseek} by 25.9 percentage points. On TreatmentPC, \name achieves 82.9\% accuracy, exceeding GPT-5 by 10.7 percentage points and DeepSeek-R1 by 15.4 percentage points. \name generalizes across brand names, generic names and diverse drug descriptions (BrandPC, GenericPC and DescriptionPC benchmarks; \extfigref{fig:descriptionpc}; Supplementary Note~\ref{sec:note1_si}).

We evaluate \name in three real-world settings. First, experts from 28 rare disease organizations assessed blinded responses to rare disease treatment cases spanning neurodevelopmental disorders, epilepsies, metabolic diseases, rare cancers, channelopathies, and immune-mediated diseases, and preferred \name over reference models across all eight evaluation criteria, with the largest gains in cognitive traceability and helpfulness of rationale. Second, practicing physicians evaluated \name on complex hospitalized patient cases in cardiovascular management and infectious disease, including post-CABG patients with CKD, anticoagulated patients with surgical site infections, and post-STEMI patients with severe asthma. Third, we tested adverse event hypotheses generated by \name in longitudinal health records from 5.4 million patients, prioritizing predictions where prior pharmacovigilance evidence was limited or absent; predicted associations reached adjusted odds ratios of 1.48--1.84 in the highest risk patient subpopulations, while negative controls remained near null.

\section*{Results}

\subsubsection*{\name reasons over treatment choices by gathering evidence step by step}

\name performs treatment reasoning by combining step-by-step analysis with access to medical evidence (Figure~\ref{fig:fig1}). It calls tools from a 212-tool biomedical library (\suptabref{tab:tool_list}) to retrieve evidence about drugs, diseases and patient populations from curated sources~\cite{kass2016openfda,ochoa2023next}. These tools support queries about indications, contraindications, drug-drug interactions, pharmacology, adverse reactions, disease phenotypes, therapeutic targets and patient-population restrictions. For example, \name can retrieve a drug's current approved indications, identify contraindications for a candidate therapy, check interactions between co-medications, map a disease to associated phenotypes, and query target phenotypic evidence. Because these tools are queried in real time, \name can incorporate current information from FDA prescribing information and biomedical knowledge bases rather than relying solely on knowledge encoded in model parameters. \name uses the retrieved evidence to guide the next reasoning step, allowing mechanisms, interactions, contraindications and safety constraints to be evaluated together.

At each step, \name determines what information is needed, selects relevant tools, retrieves evidence and incorporates the returned information into the analysis~\cite{yao2023react}. It continues this process until the evidence supports a final answer. The output includes both the answer and a reasoning trace that records which evidence was retrieved and how it was used.

\name can also break complex treatment tasks into smaller analyses. A patient scenario may require identifying candidate treatments, checking drug-drug interactions, evaluating comorbidities, comparing safety warnings and applying patient-specific constraints. \name analyzes these components and combines the results into a final answer (Figure~\ref{fig:fig1}). Additional details of the inference process are provided in~\onlineref{sec:skill_txagent} and Algorithm~\ref{alg:txagent_inference}; examples of \name's key agentic abilities (knowledge grounding, goal-oriented tool selection, multi-step reasoning and real-time retrieval) are shown in \supfigref{fig:extend_4abilities}.

\subsubsection*{Self-learning for treatment reasoning 
}

Multi-step treatment reasoning traces are too large and varied to annotate by hand~\cite{zelikman2022star,wang2023selfinstruct}. Each trace must specify what evidence to retrieve, which tools to use, how to interpret the returned information and how to combine evidence across multiple reasoning steps. \name is therefore trained through two levels of self-learning that replace human written traces with generated reasoning trajectories~\cite{ouyang2022training}. The first level teaches \name the structure of treatment reasoning, including problem decomposition, evidence retrieval, tool use and evidence interpretation. The second level teaches \name how to act within this structure by improving tool selection, evidence gathering and exploration of alternative reasoning paths.

At the first level, \name automatically constructs its own training data. Generating treatment reasoning traces directly would require a model that already solves the task. Instead, a collection of agent systems generates biomedical tools, treatment tasks and multi-step reasoning traces. This process produces \trainset, a dataset of 378,027 instruction tuning samples derived from 85,340 treatment tasks, comprising 177,626 reasoning steps and 281,695 tool calls grounded in FDA drug labels since 1939 (\extfigref{fig:multi-agent-system}). Supervised fine-tuning on \trainset yields the initial \name model.

At the second level, \name refines its policy through reinforcement learning (\extfigref{fig:rl-system}). During training, \name explores the 212 biomedical tools used at inference and generates multi-turn reasoning trajectories for each prompt. Each trajectory receives scientific feedback based on rewards for answer correctness~\cite{guo2025deepseek}, output format validity~\cite{qian2025toolrl}, evidence gathering, multi-step reasoning, tool-argument grounding and reasoning non-redundancy. Group relative policy optimization~\cite{shao2024deepseekmath,zhao2025geometric} then increases the probability of higher-scoring trajectories. This process improves tool selection and evidence gathering across reasoning steps. Training details are provided in \onlineref{sec:train_set}.

\subsubsection*{\name outperforms language models on Drug Prescribing Card reasoning}

We evaluate \name on DrugPC (Drug Prescribing Cards), a dataset of 3,168 treatment questions (Figure~\ref{fig:fig2}a). DrugPC covers 11 tasks of drug information, including drug overview, ingredients, warnings and safety, dependence and abuse, dosage and administration, use in specific populations, pharmacology, clinical information, nonclinical toxicology, patient information and storage and supply. To reduce data leakage from pretraining~\cite{golchin2024timetravel}, the evaluation uses drugs approved by the FDA in 2024, while excluding these drugs from training (Methods~\ref{sec:benchmark_details}).

We evaluate models in an open-ended setting~\cite{singhal2025toward}. Each treatment question is presented without answer choices. The model generates a free-form response, which is then mapped to the correct option among the original 4-5 answer choices~\cite{su2025kgarevion}. Treatment questions are verified by human experts for validity (Table~\ref{tab:example_question_type}; additional details are provided in \onlineref{sec:benchmark_details}).

We compare \name to GPT-5, DeepSeek-R1 and Qwen3 across all 11 tasks (Figure~\ref{fig:fig2}b). \name achieves 94.7\% overall accuracy across the 3,168 questions. GPT-5 achieves 76.9\%, DeepSeek-R1 achieves 68.8\% and Qwen3 achieves 48.7\%. \name improves accuracy by 17.8 percentage points over GPT-5, 25.9 percentage points over DeepSeek-R1 and 46.0 percentage points over Qwen3. Performance remains consistently high across categories, including warnings and safety, dosage and administration, and use in specific populations.
These results show that drug reasoning benefits from iterative evidence gathering and tool use. Rather than relying solely on parametric knowledge, \name retrieves and interprets FDA label information through multi-step reasoning, improving performance on tasks that require integrating indications, dosing, safety information and population-specific treatment constraints.
Ablations isolating the contribution of multi-step reasoning, training trace length, inference step budget, tool execution, and tool library scaling are reported in \extfigref{fig:avg_infer_step} and Supplementary Note~\ref{sec:note3_si}.

\subsubsection*{\name outperforms language models on patient treatment selection}

We evaluate \name on TreatmentPC, a dataset of 456 patient-specific treatment cases. Each case compares drugs indicated for the same disease but differing in properties relevant to treatment selection, including indications, use in specific populations, safety warnings, precautions, contraindications and drug interactions. The correct answer depends on patient context. A treatment may be appropriate for the disease but unsuitable because of pregnancy, comorbidity, dosing constraints or a contraindicated co-medication~\cite{huang2024foundation}. We evaluate models in an open-ended setting~\cite{singhal2025toward,su2025kgarevion}, where the model generates a free form response that is subsequently mapped to one of the answer options (\onlineref{sec:benchmark_details}).

\name achieves 82.9\% accuracy on TreatmentPC, outperforming all reference models (Figure~\ref{fig:fig2}c). \name exceeds GPT-5 by 10.7 percentage points, DeepSeek-R1 by 15.4 percentage points, Qwen3-Next by 22.8 percentage points and Qwen3 by 43.7 percentage points. Tool use LLMs with access to \toolbox perform substantially worse. ToolACE-8B~\cite{liu2024toolace} achieves 13.4\% accuracy and WattTool-8B~\cite{watt-tool-8B} achieves 5.9\%.

These results reveal a central challenge of treatment reasoning: access to biomedical tools alone is insufficient. Models must determine what evidence is needed, select relevant tools, interpret returned information and revise conclusions when evidence is incomplete, conflicting or unexpected.

When GPT-5 is given optional access to \toolbox, it invokes tools on only 1\% of treatment cases and its accuracy falls below its own no-tool baseline. When tool use is required, performance does not recover. In contrast, \name invokes tools on every treatment case and integrates retrieved evidence into subsequent reasoning steps, achieving 82.9\% accuracy. The limiting factor in LLM treatment reasoning is therefore not tool availability but the learned capacity to reason over tool outputs, and providing a frontier model with direct access to \toolbox does not substitute for that capacity (Figure~\ref{fig:fig2}d; Supplementary Note~\ref{sec:note3_si}).

TreatmentPC requires joint reasoning over patient case and drug properties. DeepSeek-R1~\cite{guo2025deepseek} and GPT-5~\cite{singh2025openai} are designed for long chain-of-thought reasoning and test-time scaling. To enable multi-step reasoning in DeepSeek-R1, we prompt it using \textless think\textgreater{} and \textless\textbackslash think\textgreater{}. Despite DeepSeek-R1's 671 billion parameters, \name outperforms it by 15.4 percentage points (82.9\% versus 67.5\%). Unlike models that rely primarily on internal knowledge, \name retrieves FDA labels and drug annotations before answering, allowing each conclusion to be grounded in source evidence. \extfigref{fig:comparison_deepseek} illustrates this on a pediatric corticosteroid scenario in which DeepSeek-R1 incorrectly judges the drug as safe, whereas \name retrieves the FDA label and identifies HPA axis suppression as a documented pediatric risk~\cite{kenalog2018label,ahmet2019adrenal}.

Both levels of \name training contribute to TreatmentPC performance (Figure~\ref{fig:fig2}e). In this analysis, the model first generates a free-form reasoning trace and answer, and then selects the answer option that best matches its own response (the ``self as judge'' protocol; Figure~\ref{fig:fig2}d). Under this setting, the Qwen3-8B base model achieves 39.2\% accuracy. The first level of \name self-learning, supervised fine-tuning on \trainset, increases accuracy to 66.5\% (+27.3 percentage points). The second level of \name self learning, reinforcement learning with scientific feedback, further improves accuracy to 74.8\% (+8.3 percentage points).

The choice of answer-extraction protocol affects absolute scores but not the direction of the result. Under the ``self as judge'' protocol, \name achieves 74.8\% accuracy; under the ``GPT-5 as judge'' protocol used for all baselines, \name achieves 82.9\%. The 8.1 percentage point difference confirms that extraction protocol contributes to absolute performance, but \name's ``self as judge'' score of 74.8\% still exceeds GPT-5's 72.2\% obtained under the more favorable ``GPT-5 as judge'' protocol, indicating that \name's advantage over GPT-5 holds regardless of which answer extraction method is applied.

\subsubsection*{Disease experts prefer \name for rare disease treatment reasoning}

We evaluated \name with disease experts from the Chan Zuckerberg Initiative Rare As One network, which brings together a network of rare disease organizations. Rare diseases often lack treatment pathways~\cite{tambuyzer2020therapies}, requiring treatment decisions to be made from limited and fragmented information about disease mechanisms, contraindications, drug interactions and rare disease patient  risks~\cite{boycott2018rare}. Experts from participating organizations provided disease-specific context used to construct treatment cases spanning neurodevelopmental disorders, epilepsies, metabolic diseases, rare cancers, channelopathies and immune-mediated diseases. 
The evaluation involved 29 experts from 28 disease organizations (\suptabref{tab:thera_consortium}), including clinicians, disease researchers and patient advocates with expertise spanning therapeutic development and clinical care. In total, 23 evaluators completed blinded assessments, contributing 110 expert evaluated responses.

\name and reference models generated responses with full reasoning traces. The primary reference model was Qwen3-8B, the base model from which \name is built, chosen to isolate the contribution of \name's tool library and self learning while holding the underlying language model fixed. Additional comparisons to o3-mini, Gemini-2.0-Flash and DeepSeek-R1 variants (full list in \onlineref{sec:human_eval}) yielded consistent preferences for \name, though the smaller sample per model precludes formal statistical comparison. Experts evaluated paired outputs in a blinded, arena based setting (Figure~\ref{fig:fig3}a,b)~\cite{chiang2024chatbot,singhal2025toward}. For each treatment case, evaluators viewed two responses without model identity and provided pairwise preferences and absolute ratings across eight criteria: task success, helpfulness of rationale, cognitive traceability, possibility of harm, alignment with clinical consensus, accuracy of content, completeness and clinical relevance (\onlineref{sec:human_eval}, Supplementary Note~\ref{sec:note5_si})~\cite{singhal2023large,pfohl2024toolbox}. Pairwise comparisons used four options: ``Model~A is better,'' ``Model~B is better,'' ``Both are equally good,'' or ``Neither did well.'' Absolute ratings used a 1--5 Likert scale and an ``Unable to Judge'' option.

Experts preferred \name over reference models across all eight criteria (Figure~\ref{fig:fig3}c). Percentages were computed over all 110 evaluations; comparison results are reported in Note~\ref{sec:note5_si} and Table~\ref{tab:stat_tests}. Preferences were strongest for cognitive traceability (95.5\%) and helpfulness of rationale (94.5\%), indicating that experts valued responses that exposed the evidence gathering and reasoning process. Across the remaining criteria, \name was preferred in 57--66\% of evaluations, whereas reference models were preferred in 16--20\%. Including ties, \name matched or exceeded reference models in 74--77\% of evaluations for completeness (66.4\% win, 10.0\% tie), task success (63.6\% win, 10.9\% tie), possibility of harm (61.8\% win, 11.8\% tie), alignment with clinical consensus (59.1\% win, 18.2\% tie), accuracy of content (58.2\% win, 18.2\% tie) and clinical relevance (57.3\% win, 18.2\% tie).

Absolute ratings confirmed that experts assigned higher quality scores to \name responses than to reference model responses. \name achieves a mean score of $4.16 \pm 0.90$ out of 5, compared with $2.44 \pm 1.26$ for reference models (Figure~\ref{fig:fig3}d). Scores are highest for cognitive traceability ($4.67 \pm 0.61$) and helpfulness of rationale ($4.58 \pm 0.69$), and lowest for possibility of harm ($3.76 \pm 0.81$). The largest gaps occur for cognitive traceability ($\Delta = 3.13$) and helpfulness of rationale ($\Delta = 2.93$), where reference models score $1.54 \pm 0.96$ and $1.65 \pm 1.09$. \name also improves accuracy of content ($3.99$ vs.\ $2.82$), completeness ($3.88$ vs.\ $2.46$) and alignment with clinical consensus ($4.18$ vs.\ $2.76$). All differences are significant (binomial test and Wilcoxon signed rank test, $P < 5 \times 10^{-5}$; \onlineref{sec:human_eval}, Supplementary Note~\ref{sec:note5_si}). 

\subsubsection*{\name reasons through complex, real-world treatment decisions}

We next evaluated \name for patient cases where treatment choices require judgment across incomplete guidelines and competing physiological constraints. We considered five clinical cases from real adult and neonatal hospitalized patients. Three physicians  independently rated \name's responses on the three cases selected for formal evaluation, spanning cardiovascular management and infectious disease, using eight criteria on a 1--5 Likert scale: task success, helpfulness of rationale, cognitive traceability, possibility of harm, alignment with clinical consensus, accuracy of content, completeness and clinical relevance. The three formally evaluated cases are shown in Figure~\ref{fig:fig4}. Two additional cases, a preoperative polypharmacy case and a neonatal case rated by a single physician, and all five reasoning traces are provided in Supplementary Note~\ref{sec:note7_si}. Case construction, rubric anchors, ethical framework and statistical treatment are described in \onlineref{sec:mount_sinai}.

On each case, \name identified the principal therapeutic risk and proposed a recommendation grounded in cited drug labels, adverse event, and drug mechanism evidence.
(1) In a post-CABG patient with CKD stage~2, recent contrast-induced nephropathy, and HFrEF~\cite{heidenreich2022aha}, \name selected enalapril over the pre-admission lisinopril on the basis of reversible effects on BUN and creatinine and the absence of specific CKD contraindications in its label, in contrast to lisinopril's more frequently reported renal dysfunction and hyperkalemia. Reviewer-level means across the eight criteria were $3.50$, $3.38$, and $3.50$, the most consistent agreement of the three cases (standard deviation across reviewers $= 0.07$; Figure~\ref{fig:fig4}a, b; Supplementary Note~\ref{sec:note7_si}, Case~1).
(2) In a patient with a mechanical mitral valve on warfarin who developed a surgical-site infection after total knee arthroplasty with gram-positive cocci on wound culture, \name flagged the levofloxacin-warfarin interaction~\cite{holbrook2005warfarin} as the decisive risk and proposed vancomycin, clindamycin, or linezolid (alongside penicillin-class beta-lactams) as alternatives effective against the cultured organism with more favorable profiles in anticoagulated patients. Reviewer level means were $2.75$, $3.63$, and $4.25$ (SD $= 0.75$; Figure~\ref{fig:fig4}c, d; Supplementary Note~\ref{sec:note7_si}, Case~2).
(3) In a post-STEMI patient with severe persistent asthma and type~2 diabetes, \name ruled out propranolol on bronchoconstriction grounds, recommended \textbeta-blockers with favorable profiles in asthma (metoprolol, bisoprolol, or carvedilol)~\cite{salpeter2002cardioselective}, and flagged hypoglycemia symptom masking as a secondary monitoring concern. Reviewer level means were $4.75$, $4.38$, and $2.17$, the highest overall per-reviewer mean of the three cases but also the widest spread (SD $= 1.40$; Figure~\ref{fig:fig4}e, f; Supplementary Note~\ref{sec:note7_si}, Case~3), with one reviewer scoring the response markedly below the other two.

Across the three cases, \name received high ratings for task success, with a mean score of $4.63 \pm 0.52$ across nine case--reviewer pairs. Every reviewer scored every case at 4 or above for this criterion. Helpfulness of rationale ($3.75 \pm 0.71$) and alignment with clinical consensus were rated above 3 by all reviewers on all cases. Across all 24 case--criterion cells, at least two of three reviewers rated \name at 3 or above. Reviewer disagreement was concentrated in cases without a single guideline endorsed treatment, consistent with the intended ambiguity of the selected scenarios. These cases illustrate how \name reasons through individual treatment decisions rather than providing a systematic quantitative evaluation; the systematic comparison is the cross-organizational expert study reported above. Per-criterion means, standard deviations and handling of reviewer-missing ratings follow the analysis plan in \onlineref{sec:mount_sinai}.

\subsubsection*{Population-scale health records support treatment-associated risk hypotheses}

We next evaluated whether \name can generate clinically meaningful hypotheses about treatment-associated adverse event risks. Many adverse events arise not from a disease, comorbidity or medication alone, but from their interaction within specific patient populations. To study this setting, we defined triadic patient cases consisting of a primary disease, a comorbidity and a medication (Figure~\ref{fig:fig5}a). For each case, \name generated candidate adverse events associated with the combined context. To focus on context specific risks, we excluded adverse events predicted from the disease, comorbidity or medication individually and retained those specific to the full triad. Candidate hypotheses were then screened against pharmacovigilance literature by a clinician and a general-purpose LLM~\cite{zheng2023judging} and prioritized for analysis where supporting evidence was limited or absent from the prior literature, making EHR evaluations a test of hypothesis generation rather than confirmation of established drug safety signals (Methods Section~\ref{sec:ehr_eval}, Supplementary Note~\ref{sec:si_ae_generation}).

We evaluated these hypotheses using cohort analyses~\cite{sherman2016rwe} in longitudinal health records from 5.4 million patients (Figure~\ref{fig:fig5}b; Methods Section~\ref{sec:ehr_eval}). For each predicted adverse event, we measured prevalence and estimated confounder adjusted odds ratios using multivariate regression. Models adjusted for age, sex, socioeconomic status and outpatient healthcare utilization. Residual confounding may remain because clinical factors, treatment history and disease severity are not fully captured by these variables~\cite{hernan2016target} (Supplementary Note~\ref{sec:note6_si}). Across disease--comorbidity--drug contexts, adverse events predicted by \name showed higher prevalence in the most specific patient subpopulations and elevated adjusted odds ratios relative to broader comparison cohorts.

Cohort analyses supported \name's prediction of increased risk of acute kidney failure in patients with hypertension and gout treated with \textbeta-blockers (OR = 1.84, 95\% CI: 1.69--2.00; Figure~\ref{fig:fig5}c, \ref{fig:fig6}a). Antihypertensive therapy has been associated with acute kidney injury (AKI), although prior analyses have not consistently identified a specific signal for \textbeta-blockers~\autocite{albasri_association_2021}. \textbeta-blockers are widely used in patients with hypertension and chronic kidney disease (CKD) and are generally considered hemodynamically neutral with respect to renal perfusion and glomerular filtration~\autocite{palmer_renal_2002}. However, \textbeta-blockers increase serum uric acid levels and are associated with incident gout~\autocite{choi_antihypertensive_2012}. Hyperuricemia is associated with renal microvascular dysfunction, inflammation and increased risk of AKI~\autocite{xu_hyperuricemia_2017}. This supports a pathway in which elevated uric acid levels in patients with gout act as an effect modifier, identifying a subgroup in which \textbeta-blocker associated metabolic effects may contribute to adverse renal outcomes. Disease severity and baseline renal function are the variables most likely to contribute to residual confounding in this analysis, as both influence \textbeta-blocker selection and independently predict renal outcomes. Clinically, AKI, including severe forms extending to acute renal failure, carries substantial morbidity and mortality, particularly in patients with multiple comorbidities.

For the same patient subpopulation, \name predicted increased risk of hyperkalemia, supported by cohort analyses (OR = 1.78, 95\% CI: 1.59--2.00; Figure~\ref{fig:fig5}d, \ref{fig:fig6}a). \textbeta-blockers have been associated with a modest but statistically significant increase in hyperkalemia risk after adjustment for comorbidities and monitoring frequency~\autocite{chang_antihypertensive_2016}. Mechanistically, \textbeta\textsubscript{2}-adrenergic blockade reduces cellular potassium uptake, impairing extrarenal potassium handling and increasing serum potassium levels~\autocite{rosa_adrenergic_1980}. As described above, \textbeta-blockers increase uric acid levels and gout risk, which is associated with AKI. Because renal excretion accounts for most potassium elimination, reduced kidney function further increases hyperkalemia risk. This defines a sequential pathway from \textbeta-blockade to hyperuricemia to AKI to impaired potassium clearance. Clinically, hyperkalemia is a life threatening electrolyte disturbance associated with arrhythmias, sudden cardiac death and interruption of guideline-directed therapies, making identification of high risk subgroups important for monitoring and prevention.

Cohort analyses supported \name's prediction of increased risk of hepatocellular carcinoma in patients with diabetes and ischemic heart disease treated with DPP-4 inhibitors (OR = 1.48, 95\% CI: 1.17--1.88; Figure~\ref{fig:fig5}e, \ref{fig:fig6}a). Diabetes is a well-established risk factor for hepatocellular carcinoma (HCC), driven by insulin resistance and chronic inflammation leading to steatohepatitis and cirrhosis~\autocite{li_diabetes_2017}. DPP-4 inhibitors are oral anti-hyperglycemic agents, although their use has declined relative to SGLT-2 inhibitors and GLP-1 receptor agonists~\autocite{do_trends_2025}. Evidence linking DPP-4 inhibitors to HCC remains inconsistent, with randomized trial meta-analyses showing no clear increase in risk compared to alternative therapies~\autocite{zhao_dipeptidyl_2017}. By contrast, GLP-1 receptor agonists and SGLT-2 inhibitors show signals of reduced HCC risk~\autocite{yang_impact_2026}. DPP-4 is involved in immune regulation and tumor biology, providing a basis for context dependent effects~\autocite{kawakita_cd26dpp-4_2021}. Ischemic heart disease likely identifies a population with greater metabolic burden and comorbidity rather than a direct interaction, although this association warrants further study.

Prevalence analyses supported \name's prediction of increased risk of squamous cell carcinoma (SCC) in patients with hypertension and gout treated with diuretics, although adjusted regression did not reach statistical significance (OR = 1.08, 95\% CI: 0.66--1.78; Figure~\ref{fig:fig5}f, \ref{fig:fig6}a). Thiazide diuretics, particularly hydrochlorothiazide, are associated with increased SCC risk in a dose-dependent manner~\autocite{schneider_risk_2021}. The mechanism likely involves photosensitization, which increases susceptibility to ultraviolet-induced DNA damage. Diuretics also increase uric acid levels and gout risk, and gout has been linked to increased cancer risk~\autocite{tian_association_2024}. These effects combine to produce a layered risk structure. Clinically, SCC can be locally invasive or metastatic, particularly in older patients with comorbidities. The absence of a significant adjusted odds ratio does not rule out a true association; SCC is a relatively rare outcome in observational data and the analysis is likely underpowered to detect modest effects. This result illustrates the range of confidence that should be applied to model generated hypotheses and motivates prospective validation in larger cohorts.

Prevalence analyses supported \name's prediction of increased risk of liver failure in patients with hyperlipidemia and hypothyroidism treated with statins, although adjusted regression did not reach statistical significance (OR = 1.04, 95\% CI: 0.68--1.58; Figure~\ref{fig:fig5}g, \ref{fig:fig6}a). Statins can cause liver enzyme elevations, but true liver failure is rare, with rates below 2 per 1 million patient years~\autocite{newman_statin_2019}. Mechanistically, statins can induce hepatocellular injury through mitochondrial and metabolic effects. Hypothyroidism is associated with steatotic liver disease and cirrhosis, which may increase baseline hepatic vulnerability. This suggests a context in which statin exposure contributes to more severe hepatic outcomes in a susceptible subgroup.

\name predicted increased risk of respiratory failure in patients with diabetes and chronic kidney disease treated with metformin; cohort analyses did not show a significant increase (OR = 1.00, 95\% CI: 0.92--1.07; Figure~\ref{fig:fig5}h, \ref{fig:fig6}a). Metformin is associated with lactic acidosis in patients with impaired renal function. Reduced clearance increases the risk of drug accumulation and acidosis. Metabolic acidosis increases respiratory drive and can contribute to respiratory decompensation~\autocite{yen_respiratory_2020}. Observational studies also report increased respiratory complications in high-risk populations with COPD. These mechanisms support indirect pathways linking metformin to respiratory outcomes in specific patient groups.

We validate the analysis using positive and negative controls. Positive controls recover established clinical effects, including hyperkalemia from ACE inhibitor therapy in CKD~\autocite{weinberg_risk_2009}, ischemic risk from inhaled \textbeta-agonists~\autocite{au_association_2002}, and acidosis from metformin in CKD~\autocite{defronzo_metformin-associated_2016}. We also recover the cardiovascular benefits of GLP-1RA and SGLT-2 therapies~\autocite{mcguire_oral_2025, pop-busui_oral_2026, natale_sodium-glucose_2024} (Figure~\ref{fig:fig6}b). Negative controls show no increase in adjusted risk, with estimates centered near null (Figure~\ref{fig:fig6}c). These results indicate that the evaluation approach does not generate spurious associations in the absence of plausible mechanisms and supports its use for evaluating plausibility of clinically meaningful adverse event risk.

\section*{Discussion}

\name is trained to perform treatment reasoning by gathering evidence through tool use, interpreting returned information, and revising its analysis over multiple steps. This process produces both a treatment recommendation and a reasoning trace that records how evidence contributes to the final conclusion. 
\name addresses treatment reasoning tasks in which a treatment case is posed and evidence must be gathered before a conclusion can be reached, tasks that arise across drug selection, personalized treatment planning, and hypothesis generation about treatment-associated risks. It is not a point-of-care reference tool or a risk calculator, but a research system to evaluate whether AI can learn the evidence seeking process that underlies treatment decisions.
Across benchmark datasets, expert evaluations, clinical cases, and population-scale analyses, \name outperformed language models and tool-use systems. Benchmark evaluations establish performance under controlled conditions, while expert evaluations in rare disease settings, physician ratings on complex hospitalized patient cases, and population-scale EHR validation test whether that performance extends to open-ended cases with real-world clinical complexity. 

A central design choice in \name is the separation of reasoning from knowledge storage. Rather than relying solely on knowledge encoded in model parameters, \name retrieves evidence from tools that query biomedical resources, including FDA drug labels and clinically validated knowledge bases, allowing conclusions to be grounded in retrievable evidence and updated as new information becomes available. This separation also has implications for how model scale relates to reasoning performance. Consistent performance of \name above DeepSeek-R1 (671B parameters) and GPT-5 suggests that targeted tool-use training may be more effective than scaling alone for tasks that require iterative evidence gathering. Together, these results show that treatment reasoning can be framed as an iterative process of evidence gathering and analysis, and that reinforcement learning can train AI to perform this process.

Several limitations warrant further investigation. The quality of \name's outputs depends on the coverage and reliability of \toolbox. Missing endpoints and retrieval errors can propagate through treatment reasoning~\cite{hager2024evaluation}. \name also does not quantify uncertainty. Although grounding in retrieved evidence improves verifiability (\extfigref{fig:comparison_deepseek}), uncertainty can arise from incomplete evidence, incorrect tool outputs, and errors in evidence synthesis~\cite{ghassemi2021false}. In the clinical case evaluations, reviewer disagreement was highest on cases without clear guideline endorsed answers, precisely the cases where quantified uncertainty would be most valuable~\cite{ouyang2022training}. In addition, training relies on generated reasoning traces rather than human written demonstrations. This enables supervision at a scale that would otherwise be impractical, but generated traces may inherit biases from the generation process~\cite{shumailov2024collapse}.

\name operates on natural language inputs and does not incorporate imaging, laboratory time series, genomic measurements, or longitudinal patient records. Extending treatment reasoning to multimodal patient data will require new representations and evaluation frameworks~\cite{moor2023foundation}. Population-scale analyses presented here to evaluate hypotheses generated by \name are observational. Although these analyses adjust for patient demographics and healthcare utilization  (Supplementary Note~\ref{sec:note6_si}), residual confounding and measurement error may remain. These results should be interpreted as support for hypothesis generation rather than causal inference.

\name frames treatment reasoning as an evidence seeking process in which AI gathers and evaluates biomedical information before reaching a conclusion. By separating treatment reasoning from knowledge storage, \name can incorporate new evidence while preserving a transparent record of how conclusions are formed. Results presented here establish a principle that treatment reasoning, which requires knowing what evidence to seek before a recommendation can be formed, can be learned through reinforcement learning over a universe of biomedical tools. As biomedical knowledge expands to encompass genomic, imaging, and longitudinal patient data, and as tool libraries expand to query these sources, training frameworks of this kind may support reasoning across an increasingly broad range of therapeutic contexts.

\clearpage
\end{spacing}

\begin{spacing}{1}
\xhdr{Acknowledgements}
We gratefully acknowledge the support of NSF CAREER 2339524, ARPA-H Biomedical Data Fabric (BDF) Toolbox Program, Harvard Data Science Initiative, Amazon Faculty Research, Google Research Scholar Program, AstraZeneca Research, Roche Alliance with Distinguished Scientists (ROADS) Program, Sanofi iDEA-iTECH Award, GlaxoSmithKline Award, Boehringer Ingelheim Award, Merck Award, Optum AI Research Collaboration Award, Pfizer Research, Gates Foundation (INV-079038),  Chan Zuckerberg Initiative, John and Virginia Kaneb Fellowship at Harvard Medical School, Biswas Computational Biology Initiative in partnership with the Milken Institute, Collaborative Center for XDP at Massachusetts General Hospital, Harvard Medical School Dean's Innovation Fund for the Use of Artificial Intelligence, and the Kempner Institute for the Study of Natural and Artificial Intelligence at Harvard University. 
A.N. was supported by the Rhodes Scholarship. 
D.A.C. was funded by an NIHR Research Professorship (NIHR302440), a Royal Academy of Engineering Research Chair, and the InnoHK Hong Kong Centre for Cerebro-Cardiovascular Engineering, and was supported by the National Institute for Health Research Oxford Biomedical Research Centre and the Pandemic Sciences Institute at the University of Oxford.
This research was enabled by the AI Cluster at the Kempner Institute for the Study of Natural and Artificial Intelligence at Harvard University. Any opinions, findings, conclusions, or recommendations expressed in this material are those of the authors and do not necessarily reflect the views of the funders. Figures~\ref{fig:fig1}, \ref{fig:fig2}, \ref{fig:fig3}, and \ref{fig:fig5} were created, in part, using Biorender.com (see \url{https://biorender.com/zn1rz6u}).

\xhdr{Ethics approval} All parts of this study that relate to the use of Clalit Health Services data were approved by the Clalit Health Services Institutional Review Board (Helsinki) committee. Work conducted at Mount Sinai was approved by the Mount Sinai Institutional Review Board (STUDY-20-00338).

\xhdr{Data availability} 
\name is available at \url{https://athena.openscientist.ai}.

\xhdr{Code availability}
All code used in this study is available at \url{https://github.com/mims-harvard/ATHENA} under MIT license. Model weights are released at \url{https://huggingface.co/mims-harvard/ATHENA-R1-Qwen3-8B}.

\xhdr{Author contributions}
S.G. developed and implemented \name, and performed all benchmarking analyses. A.N., J.K., B.G., A.S. and J.L. developed and evaluated the clinical case scenarios. S.G. and R.Z. developed the rare disease treatment reasoning evaluation. Members of the \name Evaluation Consortium performed human evaluations in rare disease treatment reasoning. A.N., N.D. and R.B. designed and performed the patient cohort analyses using electronic health records. All authors discussed the results and contributed to the manuscript. S.G. and M.Z. designed the study, and M.Z. supervised and led the overall study.

\xhdr{Competing interests}
The authors declare no competing interests.

\end{spacing}

\clearpage

\begin{spacing}{1}
\small
\newgeometry{left=0.4in,right=0.4in}
\captionsetup{margin=0.1in}
\pagestyle{empty}

\begin{figure}[ht]
    \centering
    \includegraphics[width=1\textwidth]{FIG/FIG1.pdf}
\caption{
\textsf{\textbf{\name solves precision treatment reasoning problems by retrieving and analyzing medical evidence from a biomedical tool universe.}} For a patient treatment scenario, \name generates a treatment recommendation together with a reasoning trace that records evidence retrieval, tool use, and intermediate analyses. The example shows \name adjusting therapy for a 77-year-old man with type 2 diabetes and early chronic kidney disease (eGFR 52 mL/min) receiving metformin, an ACE inhibitor, and hydrochlorothiazide. Green nodes denote human input and clinician interjections, blue nodes reasoning steps, and orange nodes tool calls that retrieve biomedical evidence. \name does not follow a fixed sequence of operations. It adaptively determines which evidence to collect, which analyses to perform, and which questions require further investigation based on information gathered in previous steps. In this example, \name evaluates the ACE inhibitor, hydrochlorothiazide, and alternative treatment options in parallel, eliminates unsupported hypotheses (here, calcium-channel blockers show no interaction with metformin), and synthesizes the remaining evidence. A clinician interjection ("Is lactic acidosis risk significant at this eGFR?") triggers additional evidence gathering and analysis before \name generates its final recommendation. Multiple orange nodes branching from a single blue node indicate parallel tool calls within one reasoning step. Separate reasoning branches indicate concurrent analyses of different treatment considerations. This adaptive process allows \name to integrate patient characteristics, medications, contraindications, and biomedical evidence when evaluating treatment options.
}
\label{fig:fig1}
\end{figure}
\clearpage

\newgeometry{left=0.7in,right=0.7in}
\captionsetup{margin=0.1in}
\pagestyle{empty}

\begin{figure}[ht]
\centering
\includegraphics[width=0.9\textwidth]{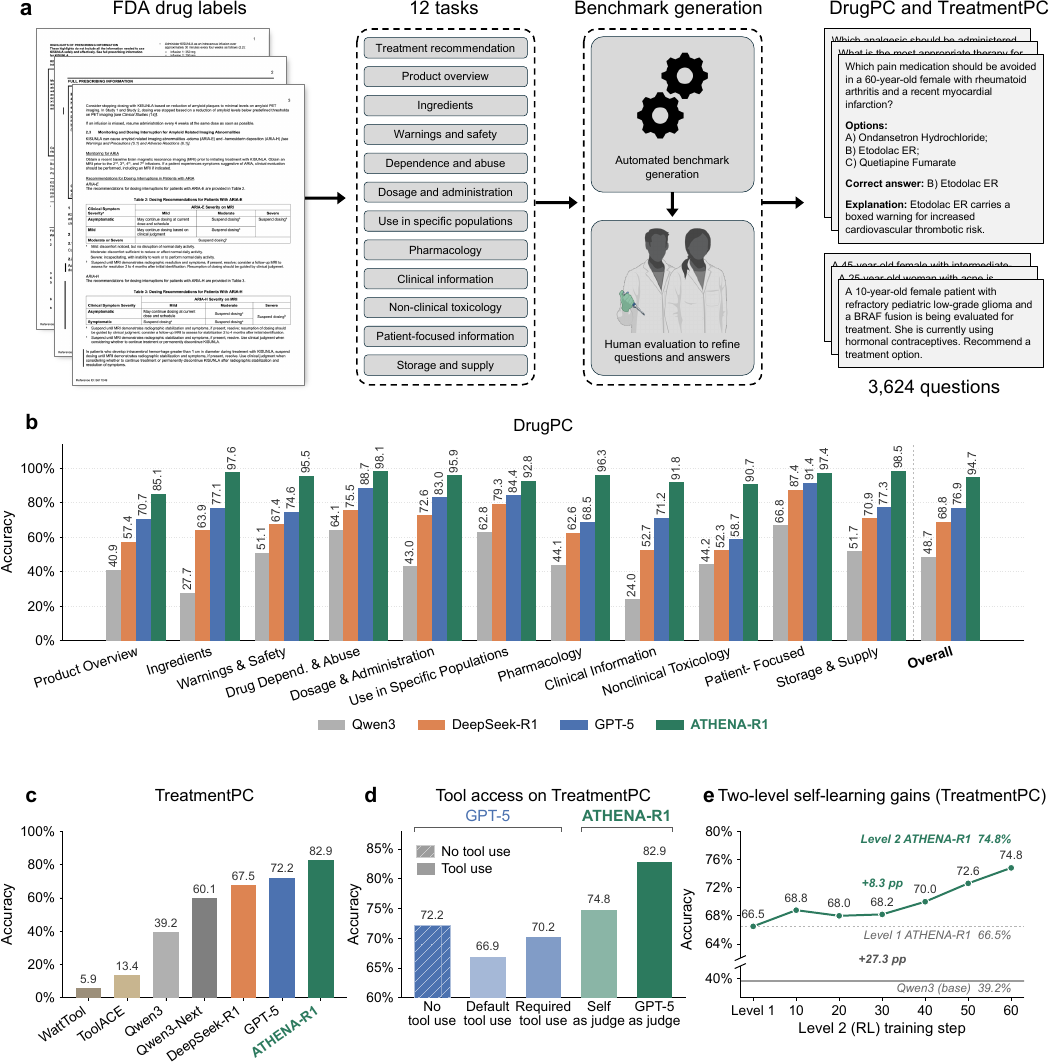}
\caption{\textsf{\textbf{\name outperforms reasoning models and tool-use LLMs on drug prescribing and patient treatment benchmarks.}}
    \textsbf{(a)} Construction of the DrugPC and TreatmentPC benchmarks from FDA prescribing information. Structured FDA drug labels were used to generate treatment questions across drug prescribing and patient-specific treatment selection tasks. Human review was used to refine the questions, answer choices, and explanations.
}
\label{fig:fig2}
\end{figure}
\clearpage

\noindent
\begin{figure}[H]\ContinuedFloat
\caption*{
    \textsbf{(b)} Accuracy on the DrugPC open-ended benchmark, comprising 3{,}168 questions across 11 drug-prescribing tasks. Each model generated a free-form answer without access to answer options; an independent GPT-5 instance then mapped the answer to one option for scoring, using the protocol shown in \textbf{d}. \name achieved 94.7\% micro-averaged accuracy across all questions, compared with 76.9\% for GPT-5, 68.8\% for DeepSeek-R1, and 48.7\% for Qwen3. The 11 tasks cover drug overview, ingredients, warnings and safety, dependence and abuse, dosage and administration, use in specific populations, pharmacology, clinical information, nonclinical toxicology, patient-focused information, and storage and supply.
    \textsbf{(c)} Accuracy on the TreatmentPC open-ended benchmark, comprising 456 patient-specific treatment scenarios. \name achieved 82.9\% accuracy, outperforming reasoning models GPT-5 (72.2\%), DeepSeek-R1 (67.5\%), Qwen3-Next (60.1\%), and Qwen3 (39.2\%), as well as tool-use LLMs with full access to \toolbox, including ToolACE-8B (13.4\%) and WattTool-8B (5.9\%).
    \textsbf{(d)} TreatmentPC accuracy under matched tool access and answerextraction conditions. GPT-5 reached 72.2\% without tools, 66.9\% when tools were available but optional, and 70.2\% when required to call a tool on every question. Under optional tool access, GPT-5 used tools on 1\% of questions, whereas \name used tools on every question. Because open-ended answers must be mapped to answer options before scoring, we compared two answer-extraction protocols for \name. In the `self as judge'' protocol, \name mapped its own free-form answer to an option and reached 74.8\% accuracy. In the `GPT-5 as judge'' protocol, an independent GPT-5 instance performed the same mapping and \name reached 82.9\% accuracy, the protocol used for all baselines in c. Both protocols score the selected option against the same gold answer key; they differ only in how answer extraction is performed.
    \textsbf{(e)} TreatmentPC accuracy across the two levels of \name self-learning, scored with the ``self as judge'' protocol from \textbf{d}. The Qwen3-8B base model achieved 39.2\% accuracy. Level~1, supervised fine-tuning on \trainset, increased accuracy to 66.5\%. Level~2, reinforcement learning with scientific feedback, further improved accuracy across RL steps 0-60, reaching 74.8\% at step~60. The y-axis is broken to show the Qwen3 baseline and the RL trajectory on the same plot. 
    Each \name accuracy in \textbf{b-e} comes from one independent rollout; sampling variability across $n{=}5$ independent rollouts is reported in \extfigref{fig:reproducibility} and Supplementary Note~\ref{sec:note8_si}.
}
\end{figure}
\clearpage

\begin{figure}[ht]
\centering
\includegraphics[width=\textwidth]{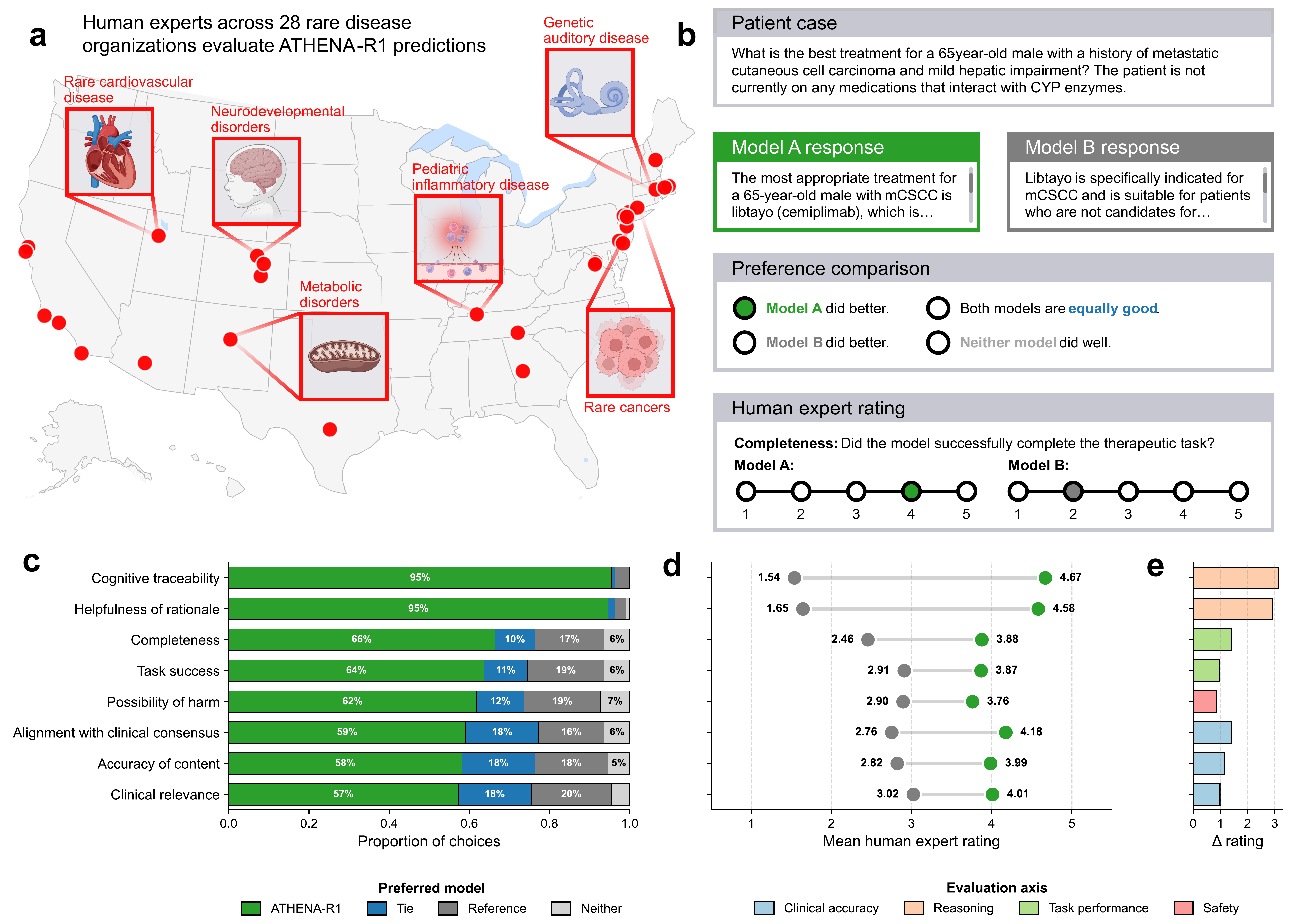}
\caption{\textsf{\textbf{Across all eight evaluation criteria, disease experts from 28 rare disease organizations prefer \name's responses and reasoning traces over those of reference models.}}
    \textsbf{(a)} Design of the human evaluation of \name.
    For each patient case and treatment-development scenario, \name and reference models (predominantly Qwen3-8B; six reference models in total, listed in Online Methods) independently generated a response and multi-step reasoning trace.
    Each case was routed to experts whose disease expertise matched the case.
    We collected 110 expert-evaluated responses from 23 evaluators, including disease experts from 28 rare disease organizations in the Chan Zuckerberg Initiative Rare As One network.
    \textsbf{(b)} Arena-based evaluation interface.
    For each treatment case, experts view two responses side by side with model identity hidden, select the preferred response (pairwise preference), and rate each response on a 1--5 scale.
    Both judgments span eight criteria: task success, helpfulness of rationale, cognitive traceability, possibility of harm, alignment with clinical consensus, accuracy of content, completeness, and clinical relevance.
    \textsbf{(c)} Pairwise preferences across all eight criteria.
    In blinded head-to-head comparisons, experts prefer \name over reference models on every criterion, with the largest margins for cognitive traceability (95.5\%) and helpfulness of rationale (94.5\%).
    \textsbf{(d)} Absolute ratings across all eight criteria.
    Experts rate \name outputs at a mean of $4.16 \pm 0.90$ out of 5, versus $2.44 \pm 1.26$ for reference models.
    \label{fig:fig3}
    \textsbf{(e)} $\Delta$ rating between \name and the reference model. Bars are colored by evaluation axis.
}
\end{figure}
\clearpage

\begin{figure}[ht]
\centering
\includegraphics[width=0.9\textwidth]{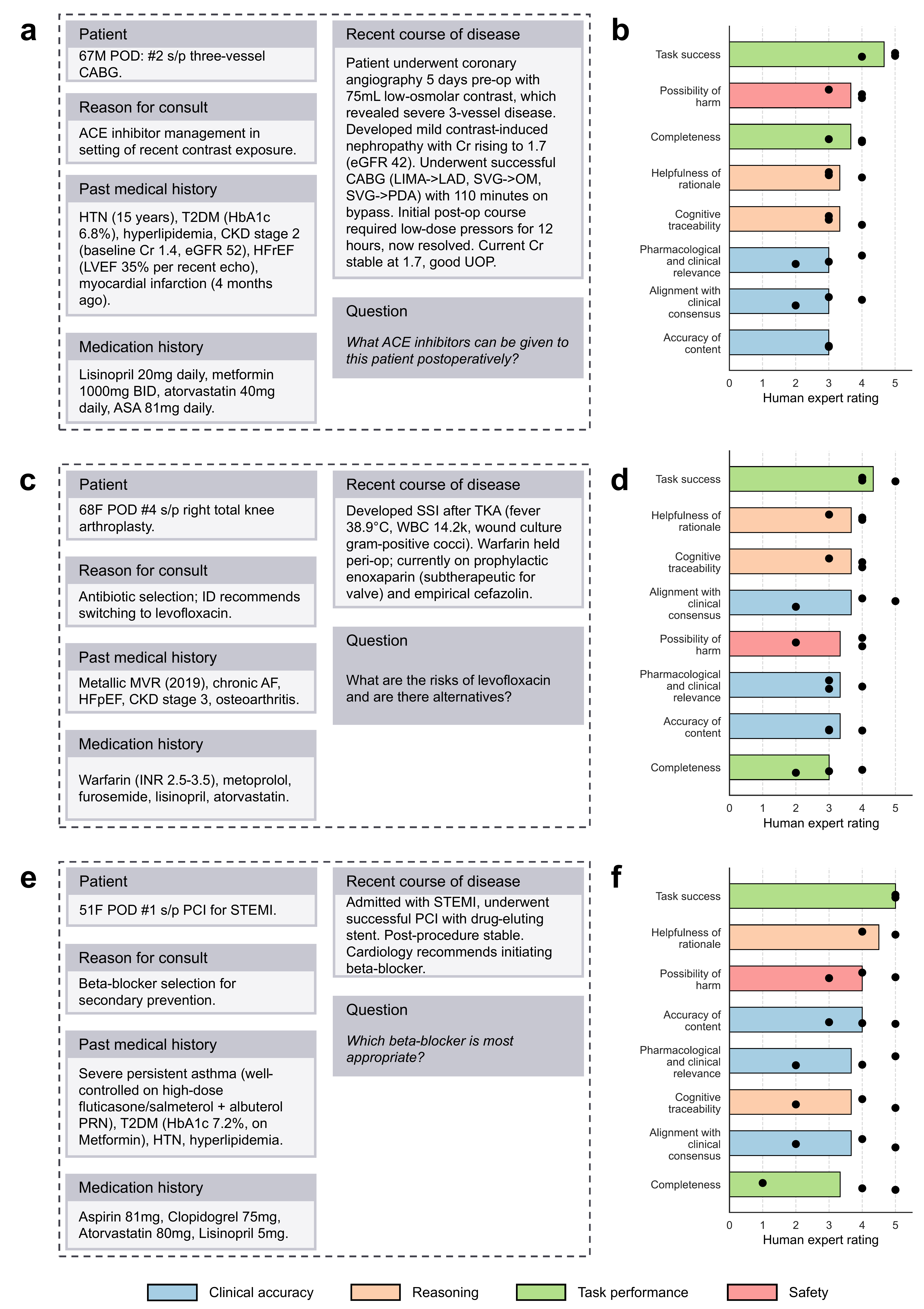}
\end{figure}

\clearpage
\captionof{figure}{\textsf{\textbf{On complex, real-world hospitalized-patient cases in cardiovascular management and infectious disease,  physician reviewers consistently rate \name's treatment recommendations as successful.
}}
    Each case is a real hospitalized patient whose treatment decision requires weighing competing physiological constraints under incomplete guideline coverage. In each case, \name identifies the principal therapeutic risk and proposes a recommendation grounded in cited label, adverse-event, and mechanism evidence. Panels pair each case (left) with its expert ratings (right).
    \textsbf{(a)} A 67-year-old male on postoperative day 2 following three-vessel CABG, with HFrEF, CKD stage 2 and recent contrast-induced nephropathy, asking which ACE inhibitors can be safely administered postoperatively.
    \textsbf{(b)} Expert absolute ratings (1--5 scale) on the case in~\textbf{a} across eight criteria: task success, possibility of harm, completeness, helpfulness of rationale, cognitive traceability, clinical relevance, alignment with clinical consensus, and accuracy of content. Bars show the mean across three physician reviewers; dots show individual reviewer scores.
    \textsbf{(c)} A 68-year-old woman with a mechanical mitral valve on warfarin, 4 days after elective total knee arthroplasty, developed a surgical-site infection with gram-positive cocci on wound culture; the infectious disease team recommended switching to levofloxacin. Question: what are the risks of levofloxacin and what are the alternative antibiotics?
    \textsbf{(d)} Expert absolute ratings on the case in~\textbf{c}, across the same eight criteria; bars and dots as in~\textbf{b}.
    \textsbf{(e)} A 51-year-old female on postoperative day 1 after PCI for STEMI with severe persistent asthma and type 2 diabetes, asking which $\beta$-blocker is most appropriate for secondary prevention.
    \textsbf{(f)} Expert absolute ratings on the case in~\textbf{e}, across the same eight criteria; bars and dots as in~\textbf{b}.
    Task success received a mean rating of $4.63 \pm 0.52$ across the nine case-reviewer pairs, and across all three cases and all eight criteria at least two of three reviewers rated \name's response at 3 or above. These cases illustrate \name's reasoning on individual treatment decisions and complement the systematic expert evaluation in Figure~\ref{fig:fig3}. Bars are colored by evaluation axis.
}
\label{fig:fig4}
\clearpage

\begin{figure}[ht]
\centering
\includegraphics[width=\textwidth]{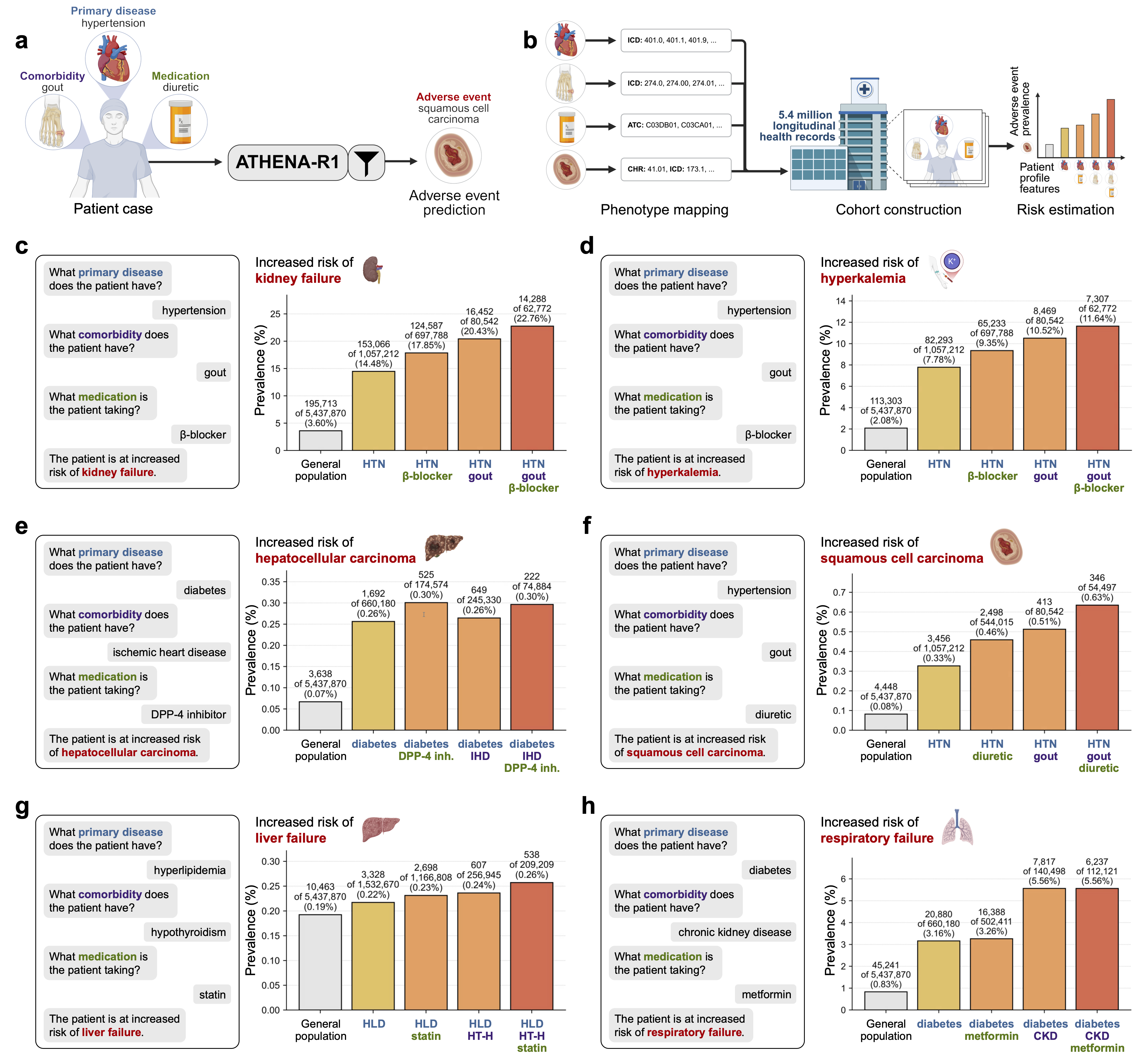}
\caption{\textsf{\textbf{Adverse events predicted by \name for disease, comorbidity, and medication profiles occur at the highest prevalence in the most specific patient subpopulations across electronic health records from 5.4 million patients.}}
    \textsbf{(a)} Workflow for generating adverse-event hypotheses with \name. We defined patient profiles, each specified by a primary disease, a comorbidity, and a medication, and used each profile to construct a contrastive prompt that directs \name to predict adverse events attributable to the full combination rather than to any single component. Candidate predictions were then scored and ranked by a clinician and a large language model.
    \textsbf{(b)} Retrospective validation pipeline using population-scale EHRs. Clinical entities (diseases, comorbidities, drugs, and predicted adverse events) are mapped to standardized medical codes. We construct patient cohorts by identifying individuals who meet the specified clinical criteria, with index dates set by the timing of diagnoses and drug exposures. Statistical analyses then quantify the risk of each predicted adverse event in exposed versus unexposed populations, using prevalence calculations and confounder-adjusted logistic regression.
    \textsbf{(c-h)} Evaluation of six adverse-event hypotheses generated by \name across patient populations drawn from electronic health records of more than 5.4 million individuals in the Clalit Health Services. Each panel shows the prompt provided to \name (left), the predicted adverse event (center), and the observed prevalence of that event across five progressively more specific patient cohorts (right).
    In all six panels, prevalence is highest in the most specific cohort, indicating that \name identifies adverse events that are most prevalent in narrowly defined patient subpopulations. Confounder-adjusted effect estimates and control analyses for these hypotheses are shown in Figure~\ref{fig:fig6}.
}
\label{fig:fig5}
\end{figure}
\clearpage

\begin{figure}[ht]
\centering
\includegraphics[width=\textwidth]{FIG/FIG6.pdf}
\caption{\textsf{\textbf{Population-scale electronic health records support \name's adverse-event risk predictions.}}
    Forest plot of adjusted odds ratios (OR) with 95\% confidence intervals (CI) for each adverse event predicted by \name, shown alongside positive controls and negative controls (red). Positive controls are established drug-risk associations that a calibrated pipeline should recover, and negative controls are biologically implausible associations that it should not. All regression models were adjusted for demographic confounders, including age, sex, and socioeconomic status. An OR above 1 whose CI excludes 1 denotes a statistically significant increase in risk.
}
\label{fig:fig6}
\end{figure}
\clearpage

\noindent
\begin{figure}[H]\ContinuedFloat
  \caption*{
     \textsbf{(a)} \name predictions: adverse-event risks predicted by \name for patient profiles defined by a disease, a comorbidity, and a medication. Three of six predicted associations reach statistical significance (OR $> 1$, 95\% CI excluding $1$): acute kidney failure (OR 1.84; 95\% CI 1.69--2.00), hyperkalemia (OR 1.78; 95\% CI 1.59--2.00), and hepatocellular carcinoma (OR 1.48; 95\% CI 1.17--1.88). All six have OR $\geq 1$.
    \textsbf{(b)} Positive controls: established clinical associations used to benchmark the EHR analysis pipeline, including known risks such as hyperkalemia from ACE inhibitor use in chronic kidney disease (OR 1.59) and known protective effects such as reduced heart failure risk with SGLT-2 inhibitors (OR 0.57).
    \textsbf{(c)} Negative controls: associations between the target drugs and unrelated medical conditions. In all cases, OR $= 1$, with confidence intervals crossing the vertical line of no effect. 
    
  }
\end{figure}
\clearpage

\restoregeometry %
\end{spacing}

\clearpage

\section*{References}
\vspace{1em}
\begin{spacing}{1}
\printbibliography[heading=none]
\end{spacing}

\end{document}